\documentclass[openany]{now}

\usepackage{epsfig}
\usepackage{graphicx}
\usepackage{amsmath}
\usepackage{amssymb}
\usepackage{caption}
\usepackage{subcaption}
\usepackage{multirow}
\usepackage{floatrow}
\usepackage{color}

\usepackage{bbm}
\usepackage{array, tabularx}
\usepackage{tabulary}
\usepackage{wrapfig}
\usepackage{appendix}
\usepackage{natbib}

\newcommand{\amt}{Mechanical Turk{ }}

\title{Crowdsourcing in Computer Vision}
\author{Adriana Kovashka \\ University of Pittsburgh \\ kovashka@cs.pitt.edu 
\and
Olga Russakovsky \\ Carnegie Mellon University \\ olgarus@cmu.edu 
\and
Li Fei-Fei \\ Stanford University \\ feifeili@cs.stanford.edu
\and
Kristen Grauman \\ University of Texas at Austin \\ grauman@cs.utexas.edu 
}

\begin{document}
\copyrightowner{A.~Kovashka, O.~Russakovsky, L.~Fei-Fei and K.~ Grauman}
\volume{10}
\issue{3}
\pubyear{2014}
\copyrightyear{2016}
\isbn{978-1-68083-212-9}
\doi{10.1561/0600000073}
\firstpage{177}
\lastpage{243} 



\frontmatter  

\maketitle

\tableofcontents

\mainmatter

\begin{abstract}
\setcounter{page}{1}
Computer vision systems require large amounts of manually annotated data to properly learn challenging visual concepts. Crowdsourcing platforms offer an inexpensive method to capture human knowledge and understanding, for a vast number of visual perception tasks. In this survey, we describe the types of annotations computer vision researchers have collected using crowdsourcing, and how they have ensured that this data is of high quality while annotation effort is minimized. We begin by discussing data collection on both classic (e.g., object recognition) and recent (e.g., visual story-telling) vision tasks. We then summarize key design decisions for creating effective data collection interfaces and workflows, and present strategies for intelligently selecting the most important data instances to annotate. Finally, we conclude with some thoughts on the future of crowdsourcing in computer vision.
\end{abstract}

\chapter{Introduction}

Data has played a critical role in all major advancements of artificial intelligence for the past several decades. In computer vision, annotated benchmark datasets serve multiple purposes:
\begin{itemize}
    \item to focus the efforts of the community on the next concrete stepping stone towards developing visual intelligence;
    \item to evaluate progress and quantitatively analyze the relative merits of different algorithms;
    \item to provide training data for learning statistical properties of the visual world.
\end{itemize}
We rely on \emph{big data} to move computer vision forward; in fact, we rely on big \emph{manually labeled} data. Harnessing this large-scale labeled visual data is challenging and expensive, requiring the development of new innovative techniques for data collection and annotation. This paper serves to summarize the key advances in this field.

In collecting large-scale labeled datasets for advancing computer vision, the key question is {\bf what} annotations should be collected. This includes decisions about:
\begin{itemize}
    \item the type of media: simple object-centric images, complex scene images, videos, or visual cartoons;
    \item the type of annotations: single image-level label, detailed pixel-level annotations, or temporal annotations;
    \item the scale of annotation: more images with sparse labels or fewer images with more detailed labels.
\end{itemize}
Different types of data come with different associated costs, including computer vision researcher time (formulating the desired dataset), crowdsourcing researcher time (user interface design and developing the annotation procedure) and annotator time (e.g., finding the visual media to annotate, or providing the semantic labels). There are tradeoffs to be made between the cost of data collection and the resulting benefits to the computer vision community.

There are two ways to optimize this tradeoff between data collection cost and the benefits for the community. The first way is to carefully considering {\bf how} data should be collected and annotated. In some cases annotators may not require any prior knowledge and this effort can be outsourced to an online marketplace such as Amazon Mechanical Turk\footnote{\url{http://www.mturk.com}}. As many other crowdsourcing platforms, \amt allows ``requesters'' to post small tasks to non-expert ``workers,'' for low cost per task. The overall cost can still be significant for large-scale data annotation efforts. This can be partially remedied by developing improved user interfaces and advanced crowd engineering techniques.

The second way to optimize the cost-to-benefit tradeoff is directly using existing computer vision algorithms to select {\bf which} data should be annotated. Using algorithms in the loop allows the annotation effort to focus specifically on scenarios which are challenging for current algorithms, alleviating human effort.

The rest of the survey is organized according to these three main questions: what, how, and which data should be annotated. Section~\ref{sec:what} discusses key data collection efforts, focusing on the tradeoffs that have been been made in deciding {\bf what} annotations should be collected. Section~\ref{sec:how} dives into the details of {\bf how} to most effectively collect the desired annotations. Section~\ref{sec:which} considers the question of {\bf which} data should be annotated and how data collection can be directly integrated with algorithmic development.

The goal of this survey is to provide an overview of how crowdsourcing has been used in computer vision, and to enable a computer vision researcher who has previously not collected non-expert data to devise a data collection strategy. This survey can also help researchers who focus broadly on crowdsourcing to examine how the latter has been applied in computer vision, and to improve the methods that computer vision researchers have employed in ensuring the quality and expedience of data collection. We assume that any reader has already seen at least one crowdsourced micro-task (e.g., on Amazon Mechanical Turk), and that they have a general understanding of the goals of artificial intelligence and computer vision in particular. 

We note that most data collection on Mechanical Turk and similar platforms has involved low payment (on the order of cents) for the annotators, and relatively small and often simple tasks (which require minutes to complete), so this is the type of annotation scenario that we ask the reader to imagine. However, crowdsourcing can also involve long-term and more complex interactions between the requesters and providers of the annotation work.

Crowdsourcing is a fairly recent phenomenon, so we focus on research in the past 5-10 years.   
Some of the most interesting approaches we overview involve accounting for subjective annotator judgements (Sections \ref{sec:attributes} and \ref{sec:subj}), collecting labels on visual abstractions (Section \ref{sec:cartoons}), capturing what visual content annotators perceive to be similar (Section \ref{sec:sim}), translating between annotations of different types (Section \ref{sec:levels}), grouping the labeling of many instances (Section \ref{sec:group}), phrasing data collection as a game (Section \ref{sec:games}), and interactively reducing the annotation effort (Section \ref{sec:active_learning}, \ref{sec:interact_time}).
The contributions we present are both algorithmic, in terms of novel mathematical formulations of solutions to vision problems interlaced with a human annotation effort, and design-based, in terms of accounting for human factors in the implementation and presentation of annotation requests.
\chapter{What annotations to collect}
\label{sec:what}

The computer vision tasks we want to solve motivate what annotations we want to collect. For example, developing computer vision algorithms that are able to automatically distinguish images of parks from images of living rooms requires manually annotating a large collection of images with binary scene class labels. Similarly, developing algorithms that are able to automatically sift through a large video archive and automatically find all instances of a person running requires annotating a large collection of videos with the temporal extent of human actions.

In this section we describe several computer vision tasks on images and videos, and summarize the key efforts
of collecting the corresponding annotations in each setting. Figures~\ref{fig:buildingblocks1}, \ref{fig:buildingblocks2} and \ref{fig:buildingblocks3} illustrate the key tasks. Note that here we do not distinguish between collecting annotations to be used at training or test time. In other words, the annotations described here can be used both to \emph{train} models to perform the corresponding task, and to \emph{quantitatively evaluate} what the models have learned.


\section{Visual building blocks}
\label{sec:visual_blocks}

\begin{figure}
\begin{tabular}{ccc}
Scene classification & Object classification & Object detection \\
\includegraphics[width=0.3\linewidth]{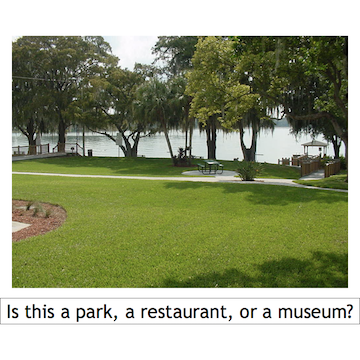}&
\includegraphics[width=0.3\linewidth]{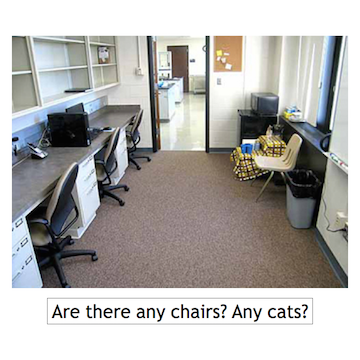} &
\includegraphics[width=0.3\linewidth]{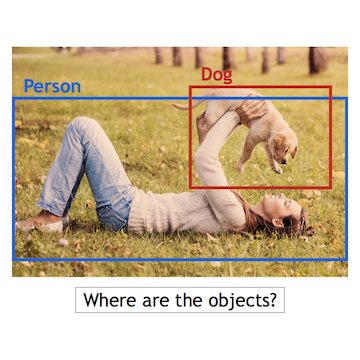}
\\
Image segmentation & Object parts & Attributes \\
\includegraphics[width=0.3\linewidth]{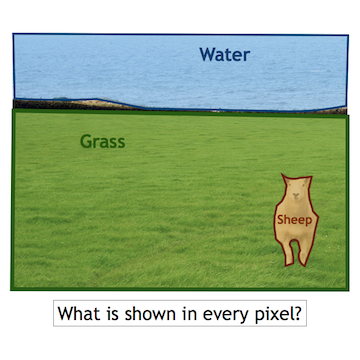} &
\includegraphics[width=0.3\linewidth]{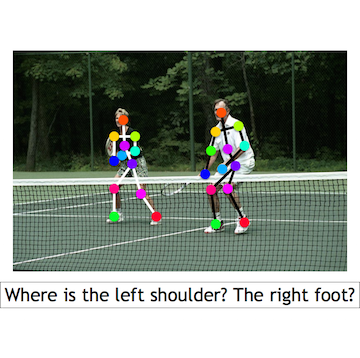} &
\includegraphics[width=0.3\linewidth]{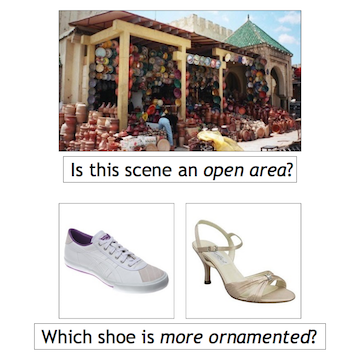}
\end{tabular}
\caption{Computer vision tasks which require the understanding of the core visual building blocks of the image. We describe strategies for collecting the corresponding annotations in Section~\ref{sec:visual_blocks}. } 
\label{fig:buildingblocks1}
\end{figure}

The most fundamental computer vision tasks require understanding the visual building blocks of an image. These tasks are illustrated in Figure~\ref{fig:buildingblocks1}. In this section we describe the challenges and the key literature related to collecting the corresponding annotations required for each of these tasks: the scene label and/or the list of objects (\emph{image classification} in Section~\ref{sec:img_cls}), the location of all the objects (\emph{object detection} in Section~\ref{sec:obj_det}), the spatial extent of all semantic regions (\emph{pixel-level image segmentation} in Section~\ref{sec:img_segm}), the spatial extent of object parts (\emph{object parts} in Section~\ref{sec:obj_parts}) and the visual properties of objects (\emph{attributes} in Section~\ref{sec:attributes}).

\subsection{Image classification}
\label{sec:img_cls}

The task of semantic image classification is arguably one of the most basic tasks in semantic image understanding. It involves assigning one (or more) class label(s) corresponding to concept(s) that appear in the image. Early efforts on this task on the datasets Caltech-101~\citep{Caltech101}, Caltech-256~\citep{Caltech256}, CIFAR~\citep{CIFAR}, 15-Scenes~\citep{Oliva01,FeiFei05,Lazebnik06} and SUN~\citep{SUN}  relied on in-house annotators to name the prominent object or the scene category in hundreds of images. However, as the desired number of visual categories and images increased, researchers had to develop effective crowdsourcing frameworks for annotation.

Most large-scale classification datasets follow a proposal/verification pipeline, where proposed images for each category are obtained from the web and then manually verified~\citep{ImageNet,Zhou14}. For example, proposed examples of ``violin'' images would be obtained by querying an image search engine for ``violin'' or semantically related queries such as ``fiddle,'' ``violin and piano'' or ``orchestra'' to obtain a large and diverse set of images~\citep{PASCALIJCV,ImageNet,ILSVRC}. The images are then manually verified by crowd workers to make sure they indeed contain the desired ``violin'' concept. This data collection strategy provides some apriori information about the concepts likely to be contained in the image, so the expensive annotation state can focus on just verifying a single concept rather than annotating an image from scratch.

In particular, the ImageNet dataset~\citep{ImageNet} contains 14,197,122
annotated images organized using the semantic hierarchy of WordNet and annotated with the help of the crowd \amt workforce using the following pipeline. Proposed images are obtained from a variety of search engines using query expansion and then manually verified through consensus; \cite{ImageNet,ILSVRC} provide a detailed overview of the design decisions. The Places dataset~\citep{Zhou14} uses a similar strategy for annotating more than 7 million images with 476 scene categories. Images are proposed using search engines and then verified in two rounds to ensure quality: first, workers are asked to select the \emph{positive} images for a given scene category, then from among the selected images other workers are asked to select the \emph{negative} images. The second verification round filters out any false positive images that may have been erroneously selected by workers in the first round. 

Image classification is not necessarily a straight-forward annotation task: e.g., determining if the object in the image is indeed a ``violin'' and not a ``viola'' may require detailed domain knowledge. Annotations that require distinguishing fine-grained categories~\citep{ImageNet,Krause13,Wah11} require an additional worker training/evaluation step to ensure that workers are qualified to perform the annotation. 


\subsection{Object detection} 
\label{sec:obj_det}

Designing computer vision models that are able to identify a single concept label per image is an important task; however, it is useful to develop systems that are able to provide a more detailed understanding of the image as well. To address this demand, tasks such as object detection were created. In object detection, algorithms are required to localize every instance of an object class with an axis-aligned bounding box. 

Multiple datasets provide axis-aligned bounding boxes around all instances of target objects. PASCAL VOC's scale of 20 target object classes and 21,738 images (training and validation set in year 2012) allowed the annotations to be done by a small group of in-house annotators~\citep{PASCALIJCV,PASCALIJCV2}. Similarly, SUN09's scale of 200 object classes and 12,000 annotated images allowed the annotations to be done by a single person~\citep{SUN}.

In contrast, the scale of ImageNet Large Scale Visual Recognition Challenge (ILSVRC) object detection~\citep{ILSVRC} makes it impossible to annotate in-house. ILSVRC annotates 1 target class in 573,966 (training and validation localization set in year 2012) and 200 target classes in 80,779 images (training and validation detection set in year 2014). An iterative crowdsourcing workflow for bounding box annotation was developed~\citep{Su12,ILSVRC}, which alternated between three steps: (1) a worker draws a bounding box around a single object instance; (2) another worker verifies the drawn box; and (3) a third worker determines if there are additional instances of the object class that need to be annotated \citep{Su12}. This system was demonstrated to be significantly more efficient than majority voting-based annotation, where multiple workers would be asked to draw a bounding box around the same object instance and their drawings would be reconciled into a single average bounding box. This is due to the fact that drawing an accurate bounding box around an object instance is several times more expensive than verifying a bounding box annotation: thus, asking one worker to draw and a few others to verify is cheaper than asking even just two workers to draw independently.

Objects are fundamental building blocks of scenes. Designing  procedures for efficiently annotating objects in large collections of images enables the collection of large-scale object detection datasets, which in turn provide benchmarks for developing and evaluating algorithms for automatic decomposition of scenes into constituent objects. 


\subsection{Pixel-level image segmentation}
\label{sec:img_segm}

\begin{figure}
\centering
\begin{tabular}{cc}
\includegraphics[width=0.48\linewidth]{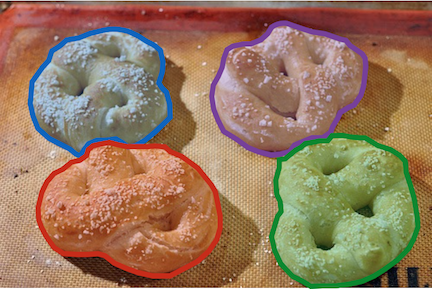} &
\includegraphics[width=0.48\linewidth]{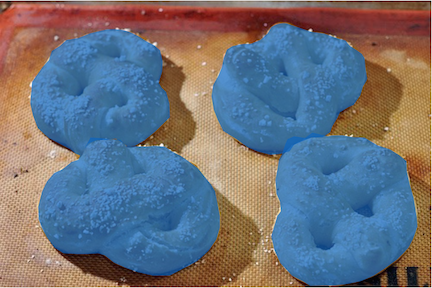}
\end{tabular}
\caption{\emph{(Left)} Instance segmentation, where each instance of ``pretzel'' is annotated separately. \emph{(Right)} Semantic segmentation, where all pixels belonging to any ``pretzel'' are annotated with a single label.}
\label{fig:segm}
\end{figure}

An additional level of complexity arises when bounding boxes are not sufficient and detailed pixel-level annotations are needed. There are roughly two types of segmentation annotations: \emph{instance-level} segmentation and \emph{semantic} segmentation, shown in Figure~\ref{fig:segm}. Instance-level segmentation datasets provide a pixel-level outline of every instance of the target objects. Semantic segmentation datasets provide an outline around contiguous regions sharing a similar semantic property. Instance-level segmentation is commonly used when annotating ``things'' (e.g., cars). Semantic segmentation is used both for annotating ``things'' (e.g., cars without distinguishing between different instances) as well as ``stuff'' (e.g., trees, sky). 

Some examples of instance-level segmentation datasets include LabelMe~\citep{Russell07}, PASCAL VOC~\citep{PASCALIJCV,PASCALIJCV2} and MS-COCO~\citep{COCO}. LabelMe~\citep{Russell07} recruits volunteers to draw polygons around object instances. However, object names and the level of detail of annotation on each image are not standardized, leading to issues with using this data to train computer vision models. PASCAL VOC~\citep{PASCALIJCV,PASCALIJCV2} uses in-house annotators to label 20 object classes in 6929 images; the Semantic Boundaries Dataset extends the annotations to five times more images from the PASCAL VOC classification and detection sets~\citep{BharathICCV2011}. The Berkeley Segmentation dataset~\citep{Arbelaez11} contains 500 images annotated with object boundaries.  The proprietary LotusHill dataset~\citep{LotusHill} contains detailed annotations of objects in 636,748 images and video frames. Among the largest instance-level image segmentation dataset to date is the COCO dataset~\citep{COCO} with more than 328,000 images with 2.5 million object instances manually segmented. 

Semantic segmentation datasets do not annotate individual object instances but do provide semantic labels for every pixel in the image. For example, SIFT Flow~\citep{Liu11} has 2,688 images labeled using the LabelMe annotation interface. Other datasets that provide pixel-level semantic segmentations include MSRC~\citep{MSRC} with 591 images and 23 classes, Stanford Background Dataset \citep{Gould09} with 715 images and 8 classes, and the PASCAL-context dataset~\citep{Mottaghi14} which annotates the PASCAL VOC 2010 images with 520 additional classes, providing a semantic label for every pixel in the image. The original PASCAL VOC 20-class instance-level segmentation dataset has also been used to evaluate semantic segmentation as well (including a 21st ``background'' class). In addition, datasets such as Weizmann Horses~\citep{Borenstein02},  CMU-Cornell iCoseg~\citep{iCoseg} or the MIT object discovery dataset~\citep{Rubinstein13} contain semantic figure-ground segmentations of a single target concept per image. 

To add even more detail to the segmentations, works such as \citep{Bell13,Bell14,Bell15,Sharan09} additionally segment and annotate material properties of images. OpenSurfaces~\citep{Bell13} provides 22,214 scenes accurately labeled with segmentations, named materials, named objects, rectified textures, reflectance judgements, and other properties. Intrinsic Images in the Wild~\citep{Bell14} follows up by annotating millions of crowdsourced pairwise comparisons of material properties. Materials in Context Database~\citep{Bell15} uses a three-stage \amt pipeline to annotate three million material samples, significantly scaling up over the previous Flickr Material Dataset~\citep{Sharan09} benchmark. These datasets enable research into deeper pixel-level image understanding.

Providing pixel-level segmentations is very time-consuming, and thus collecting segmentation datasets is particularly expensive. However, such detailed annotations enable the development and evaluation of computer vision algorithms that are able to understand the image on a much finer level than what is possible with just simple binary image-level annotations or with rough bounding box-level localization of object instances.


\subsection{Object parts}
\label{sec:obj_parts}

Besides annotating just the presence or location of objects in images, researchers have additionally looked at annotating \emph{parts} of objects. While part recognition can be evaluated as a computer vision task in its own right, more often part annotations have served to help train object recognition models by providing correspondences between different instances of the same object class. 

Multiple efforts exist to collect semantic part annotations. \cite{PASCALparts} and \cite{Azizpour12} expand the PASCAL VOC dataset of \citep{PASCALIJCV,PASCALIJCV2} by using in-house annotators to label parts of objects such as ``tail of aeroplane,'' ''beak of bird,'' or ``right eye of cow.'' Further, annotating parts or keypoints on people has been particularly popular.  \cite{Bourdev09} create a large-scale H3D dataset of human part annotations in images by designing an annotation interface that allows users to mark keypoints and displays the 3D pose in real time. These keypoints are used to discover \emph{poselets}, or groups of image patches tightly clustered in both 3D joint configuration space as well as 2D image appearance space, often corresponding to semantic parts. Follow-up work by \cite{Maji11} deploys the interface on \amt and collects human keypoint annotations on the PASCAL VOC 2010 action dataset~\citep{PASCALIJCV2}. \cite{Andriluka14} annotate position of body joints, full 3D torso and head orientation, occlusion labels for joints and body parts, and activity labels, on more than 40,522 images of people. They use pre-selected qualified \amt workers to maintain data quality, followed by manual inspection. 

Part annotations provide correspondences between different instances of the same object class. However, such correspondences do not necessarily need to be semantic. \cite{Patterson12,Deng13} and \cite{Maji13} use \amt to directly annotate spatial correspondences rather than focusing on ``nameable'' parts. \cite{Maji13} present subjects with pairs of images and ask them to click on pairs of matching points in the two instances of the category. These points may correspond to semantic structures but not necessarily. \cite{Patterson12} use \amt workers to identify clusters of image patches with strong visual and semantic similarity, providing a diverse dictionary of scene parts. \cite{Ahmed14} follow a similar strategy to obtain a dictionary of object parts. \cite{Deng13,Deng16} annotate discriminative image regions that can be used for fine-grained image classification. They do so through a ``Bubbles'' game, where \amt workers 
have to classify a blurred image into one of two object categories, by revealing only a few small circular image regions (``bubbles'').

Part annotations allow us to go beyond naming and localizing objects to understanding their spatial configuration. Localizing the parts of a bird and looking up their appearance in a field guide helps infer the bird species; understanding human pose helps infer the action the human is performing. Obtaining large-scale part annotations allows the development of computer vision models that are able to learn about and effectively utilize information about the object configuration.


\subsection{Attributes}
\label{sec:attributes}

Much of the research we have described thus far models where different objects are located, but now \emph{how they look}.
As computer vision progressed, researchers proposed a more descriptive approach to recognition, which allows visual content to be examined at a finer level than the object category level allows. 
Semantic visual attributes \citep{Lampert09, Farhadi09} were proposed as a method for describing the visual properties (often adjectives) of objects. 
For example, the same category, e.g. ``chair,'' can have different category labels: one chair might be \emph{red} and \emph{metallic}, while another is \emph{green} and \emph{wooden}. 

Collecting attribute annotations is challenging because in contrast to object categories, attributes might be perceived differently by different annotators. Further, it is not trivial to say whether an attribute is definitely present or definitely not present in the image. 

While originally attributes were modeled as binary categories \citep{ferrari2007learning, Lampert09, Farhadi09, Russakovsky10, wang2010discriminative},
\cite{parikh2011relative} proposed to model them as relative properties, where one image has the attribute \emph{more} than another. This allows the authors to obtain more reliable annotations, since for many attributes, it is more natural for a human annotator to judge an image relative to some other image, rather than in absolute terms. For example, a person who is smiling \emph{a little} (hence not definitively smiling/not smiling) might be more smiling than another person, but less than a third person.
In \citep{parikh2011relative}, a single annotator was asked to define an ordering on the \emph{categories} of images with respect to some attribute. For example, the annotator declared that Clive Owen is more masculine-looking than Hugh Laurie, who is more masculine-looking than Alex Rodriguez, etc. 
\cite{kovashka2012whittlesearch, Kovashka15b} expanded on this idea by collecting \emph{image}-level annotations in the form of pairs of images where one image has the attribute more than the other. 
The authors show this enables more accurate attribute models because it captures the variability within each category (e.g Hugh Laurie might be smiling more than Clive Owen in some images, but less in others).
\cite{yu2014fine} also collect instance-level comparisons from the crowd, but focus on very fine differences (e.g., the relative sportiness of two images which are both sporty, e.g. running shoes).
Examples of attribute annotations are shown in Figure \ref{fig:buildingblocks1}.

While collecting attribute annotations in a relative way eliminates the need to make binary decisions about attribute presence or absence, it does not solve the problem of attribute subjectivity. For example, \cite{Farhadi09} observe noticeable disagreement between annotators over attribute presence labels. \cite{Kovashka13b} find disagreement over relative labels, and propose to explicitly account for this disagreement by building individual attribute models for annotators, via adaptation from a ``generic'' model trained on data collected from the crowd. 
In a follow-up work, \cite{Kovashka15a} discover the ``shades of meaning'' of the attributes by mining the underlying latent factors behind the different label values that annotators provide. 
To gain more in-depth understanding of why annotators disagree over the labels, and as a measure of quality control, they also collect justifications from the annotators as to why they provided a particular label. For example, when asked why he/she labeled a shoe image as being ``ornate,'' a user wrote: ``The flowerprint pattern is unorthodox for a rubber boot and really stands out against the jet black background.'' Another user labeled a very similar image as ``not ornate,'' and justified it with: ``Ornate means decorated with extra items not inherent in the making of the object. This boot has a camo print as part of the object, but no additional items put on it.'' By accounting for the distinct attribute interpretations annotators employ when annotating an image, the system can build attribute models that more closely align with the internal human ``models.''

Before attribute-based applications can be developed, e.g., for object recognition \citep{Farhadi09, Branson10} or image retrieval \citep{kovashka2012whittlesearch, siddiquie2011image}, an attribute vocabulary needs to be devised, i.e., a list of attribute terms for which models will be trained. 
\cite{Patterson12} use an offline crowdsourcing approach, where they show annotators pairs of images and ask them  to list words that distinguish one image from the other. These words are then aggregated to create the attribute vocabulary. \cite{maji2012discovering} also discovers a vocabulary of attributes by asking annotators to list differences between images. 
\cite{parikh2011interactively} adopt an interactive approach to find attribute terms that are both discriminative for the machine and nameable for humans. Their method automatically proposes splits in visual space, which are then visualized for humans. Annotators are asked to provide a name for the split, or to state that the split is not nameable. The method then learns a model for ``nameability,'' which is used in selecting the future splits shown to humans for annotation. Only the terms labeled as ``nameable'' become part of the final vocabulary.
Regardless of the exact strategy used to generate attribute vocabularies, a human should be employed during some phase of the vocabulary generation process since each attribute word should be understandable by humans, if it is to be used in human-facing applications. 

 \cite{endres2010benefits} discuss the challenges of collecting annotations beyond object labels, such as attribute labels, polygons, and segmentations. They discuss issues involving the phrasing of tasks, annotator attention and how to simplify the tasks, misunderstandings by foreign language speakers, imperfect human visual processing, etc. Overall, research on attributes shows that attribute annotations are beneficial for many tasks (e.g., object recognition or image retrieval) but they require special attention as they can be ambiguous.


\section{Actions and interactions}
\label{sec:actions_interactions}

\begin{figure}
\begin{tabular}{ccc}
Actions classification & Video understanding & Cartoons \\
\includegraphics[width=0.3\linewidth]{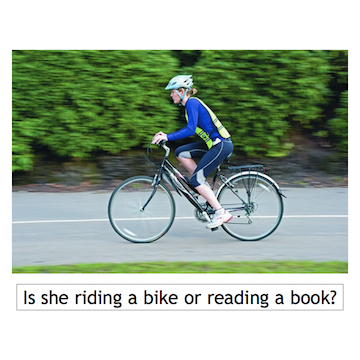} &
\includegraphics[width=0.3\linewidth]{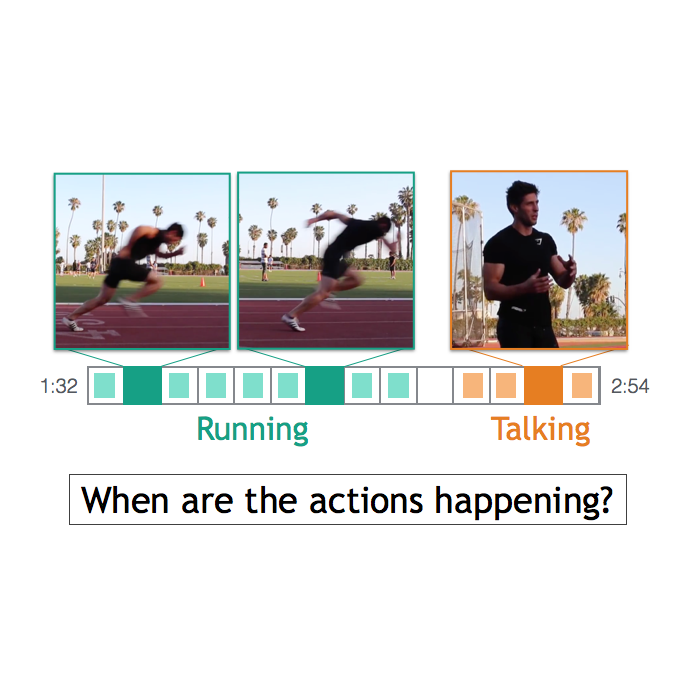} &
\includegraphics[width=0.3\linewidth]{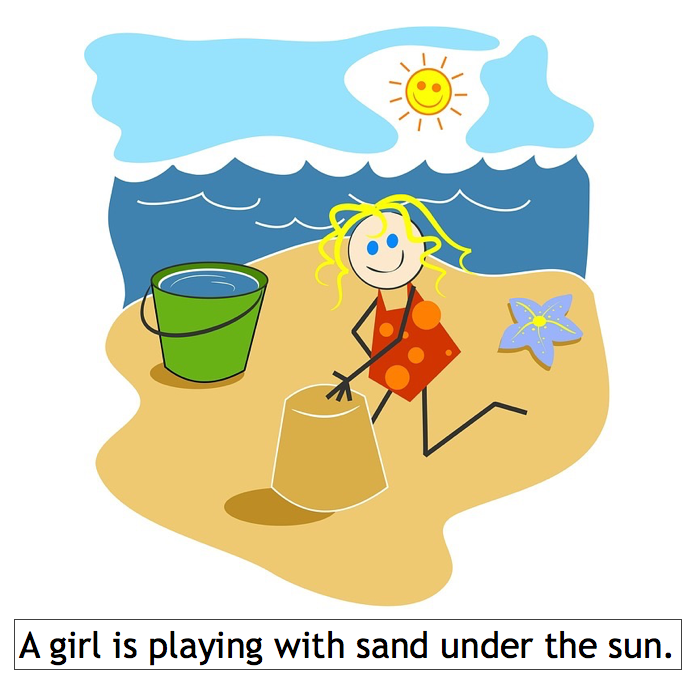}
\end{tabular}
\caption{Computer vision tasks and annotation types which shed light on the visual actions and interactions.
We describe strategies for collecting the corresponding annotations in Section~\ref{sec:actions_interactions}. } 
\label{fig:buildingblocks2}
\end{figure}

So far we discussed research into annotating the constituent components of an image: the scene label, the names and locations of objects, the spatial configuration of object parts, and the object descriptions using attributes. However, simply knowing the name and location of the different components is not enough; computer vision research aims to go deeper and understand the \emph{actions and interactions} between the different building blocks. In this section, we focus on annotating the actions and interactions that occur in images (Section~\ref{sec:actions_images}) and videos (Section~\ref{sec:actions_videos}). These types of annotations may be difficult to obtain on real-world data, so we conclude by discussing a recent line of work on using abstractions and cartoons to study interactions within a scene (Section~\ref{sec:cartoons}). Figure~\ref{fig:buildingblocks2} summarizes the tasks.

\subsection{Actions and interactions in images}
\label{sec:actions_images}

Annotating the actions and interactions between objects in images provides a more comprehensive view of the story beyond just the object/part locations. Earlier efforts such as \cite{Gupta09,PASCALIJCV,Yao10,Yao11,Le13emnlp} used in-house annotators to label 6-89 human actions (such as ``reading,'' ``riding a bike,'' ``playing guitar,'' or ``holding a guitar'').

\begin{figure}
\centering
\includegraphics[width=\linewidth]{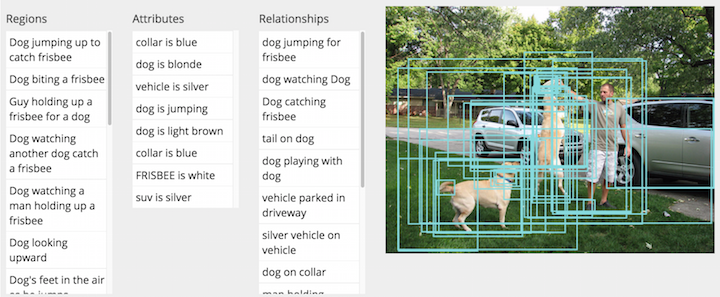}
\caption{Image annotation with objects, attributes and relationships along with spatial grounding. Visualization courtesy of \url{https://visualgenome.org}. The dataset is available from~\citep{Krishna16}.}
\label{fig:visualgenome}
\end{figure}

Larger-scale efforts of detailed annotation of human actions heavily utilize crowdsourcing. The TUHOI dataset~\citep{Le14} contains 58,808 instances of humans interacting with one or more of 200 object classes, annotated on 10,805 images from the ILSVRC object detection dataset~\citep{ILSVRC} using the CrowdFlower\footnote{\url{http://www.crowdflower.com}} service. The UT Interactee dataset~\citep{Yeh14,Yeh16} spans more than 10,000
images from SUN~\citep{SUN}, PASCAL~\citep{PASCALIJCV}, and COCO~\citep{COCO}, and contains bounding box annotations for the object (or another person) that each person in the image is interacting with. \cite{Chao15} used \amt to create the Humans Interacting with Common Objects (HICO) dataset, containing 47,774 images covering 600 categories of human-object interactions (e.g., ``ride a bike'') over 117 common actions (e.g., ``ride,'' ``feed'') performed on 80 common objects (e.g., ``bike,'' ``bear''). \cite{Johnson15} crowdsources annotation of 5,000 \emph{scene graphs} which encode objects (e.g., ``girl''), attributes (e.g., ``girl is blonde'') and relationships between objects (e.g., ``girl is holding a racket''), each grounded to regions of the image represented using bounding boxes.~\cite{Krishna16} expand this effort to annotate a large-scale Visual Genome dataset, consisting of 33,877 object categories, 68,111 attribute categories, and 42,374 relationship categories annotated across 108,077 images. Figure~\ref{fig:visualgenome} shows an example of dense image annotation with objects, attributes and relationships.\\

Some efforts go beyond simply naming human actions into developing creative ways to describe or explain human behavior. 
\cite{Vondrick16} instruct \amt workers to annotate PASCAL VOC 2012~\citep{PASCALIJCV2} and MS-COCO~\citep{COCO} images with  \emph{motivations} of actions, e.g., \emph{why} is the person sitting on the couch: Is she waiting for an appointment? Watching TV? Working on her laptop? These types of annotations of detailed interactions between different components of an image allow computer vision research to progress beyond simply localizing the people or objects towards understanding \emph{why} they are in a certain configuration in the scene. 

\subsection{Detailed video annotation}
\label{sec:actions_videos}

The best testbed for understanding the effects of actions and interactions is in the video domain rather than static images. However, 
annotating videos brings a unique set of crowdsourcing challenges. The scale of even a small video dataset can be quite large compared to images: just 10 minutes of video contains between 18K-36K frames (at 30-60 frames per second). Obtaining just video-level labels (such as ``this is a basketball video'') may only require watching a small fraction of the frames and is less expensive, but exhaustive annotation of spatio-temporal localization of objects or human actions quickly becomes prohibitively expensive.  

There is a lot of temporal redundancy between subsequent frames, allowing for obtaining annotations only on key frames and interpolating in between. Efforts such as LabelMe for video~\citep{Yuen09}, VATIC  (Video Annotation Tool from Irvine, California)~\citep{Vondrick13} or the work of \cite{Vija12} exploit temporal redundancy and present cost-effective frameworks for annotating objects in videos. The authors design interfaces for workers to label objects in a sparse set of key frames in videos, combined with either linear interpolation (in LabelMe~\citep{Yuen09}) or nonlinear tracking (in e.g., \cite{Vondrick13,Vija12}). The approaches of \cite{Vondrick11,Vija12,fathi2011combining} and others additionally incorporate active learning, where the annotation interfaces learns to query frames that, if annotated, would produce the largest expected change in the estimated object track. 

Despite these innovations, object annotation in video remains costly and scarce. The available datasets include YouTube-Objects~\citep{Prest12} with 10 object classes annotated with 6,975 bounding boxes (in version 2.2), SegTrack~\citep{Tsai10,LI13} with 24 object classes annotated with pixel-level segmentations across 976 frames total, and the ILSVRC video dataset~\citep{ILSVRC-video} with bounding boxes annotated around all instances of 30 object classes in 5354 short video snippets (by a professional annotation company using the VATIC~\citep{Vondrick13} toolbox). These spatio-temporal object tracks allow studying the physical interactions between objects and the behavior of animate objects as they move around the scene. 

The dynamic nature of videos makes them particularly well-suited for studying \emph{actions}, and thus much work has focused on annotating human actions (rather than objects) in videos. However, using crowdsourcing for large-scale video annotation remains challenging due to the size of the data and the difficulty of designing efficient interfaces. \cite{Bandla13} propose an active learning-based interface for efficient action annotation but have not utilized it for crowdsourcing. Some existing large-scale action datasets such as EventNet~\citep{EventNet} or Sports-1M~\citep{Karpathy14} rely on web tags to provide noisy video-level labels; others, like THUMOS~\citep{THUMOS} or MultiTHUMOS~\citep{Yeung15}, employ professional annotators rather than crowdsourcing to label the temporal extent of actions. 

Nevertheless, two recent large-scale video annotation efforts have successfully utilized crowdsourcing. First, ActivityNet~\citep{ActivityNet} uses a proposal/verification framework similar to that of ImageNet~\citep{ImageNet} where they define a target set of actions, query for proposal videos of those actions, and then manually clean up the results. They annotate a large dataset of 27,801 untrimmed videos with 203 human activities classes along with their temporal extent. 

Second, Hollywood in Homes~\citep{CharadesHCOMP,Charades} entirely crowdsources the creation of a video dataset, including scripting, filming and annotating videos. An \amt worker is first instructed to write a script for a 30-second video containing a few target objects and a few target actions, and another worker is then instructed to act out the script and film the video. This method has been used to create the Charades dataset of 9,850 videos showing activities of 267 workers from three continents. The dataset is then labeled with video descriptions, temporally localized actions, and object classes. 

The attention of the computer vision community is slowly shifting from understanding images towards understanding videos, and from understanding individual visual entities to understanding their actions and interactions. Going forward, efficient crowdsourcing strategies for large-scale video annotation will be critical for collecting the necessary benchmarks to advance these directions. 

\subsection{Abstraction and cartoons}
\label{sec:cartoons}

\begin{figure}
\centering
\begin{tabular}{ccc}
\includegraphics[width=0.3\linewidth]{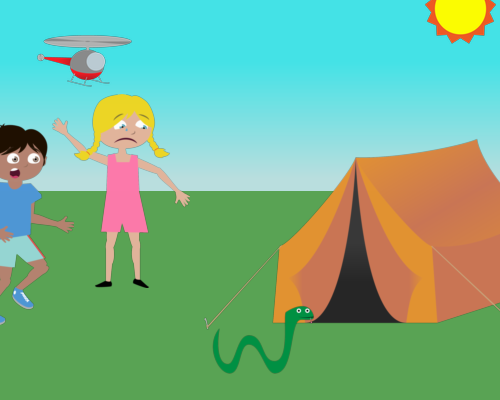} &
\includegraphics[width=0.3\linewidth]{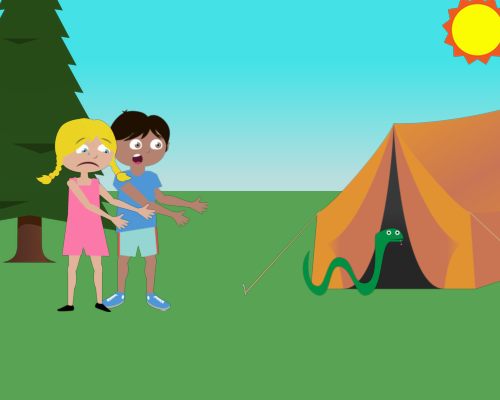} &
\includegraphics[width=0.3\linewidth]{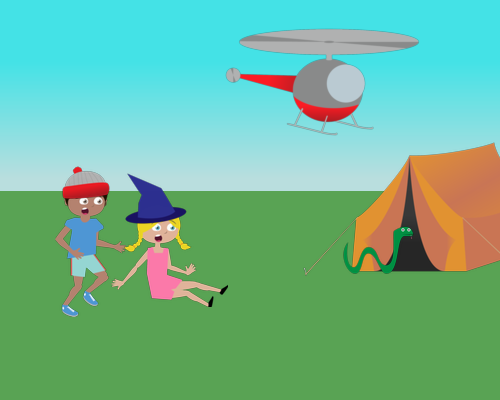} \\
\includegraphics[width=0.3\linewidth]{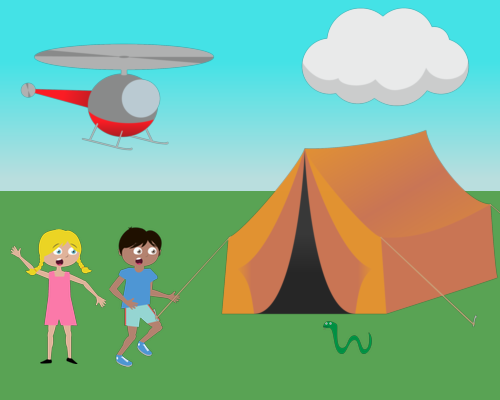} &
\includegraphics[width=0.3\linewidth]{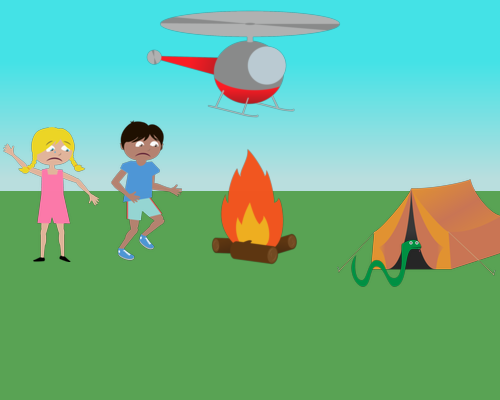} &
\includegraphics[width=0.3\linewidth]{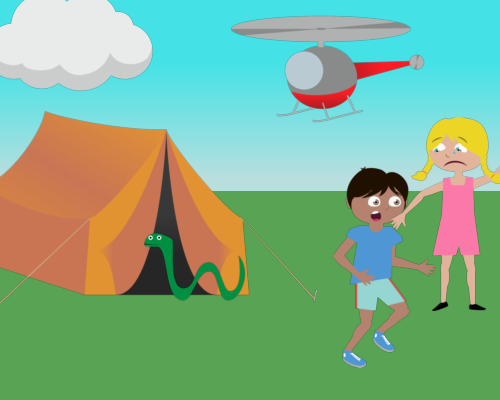} \\
\includegraphics[width=0.3\linewidth]{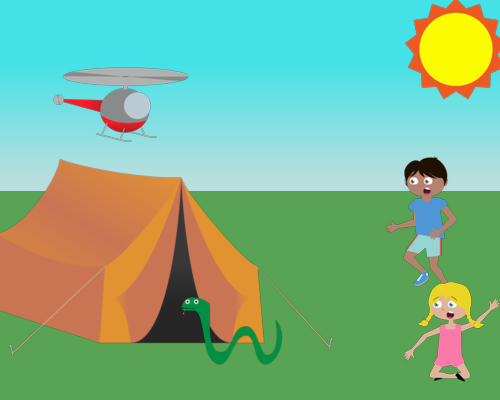} &
\includegraphics[width=0.3\linewidth]{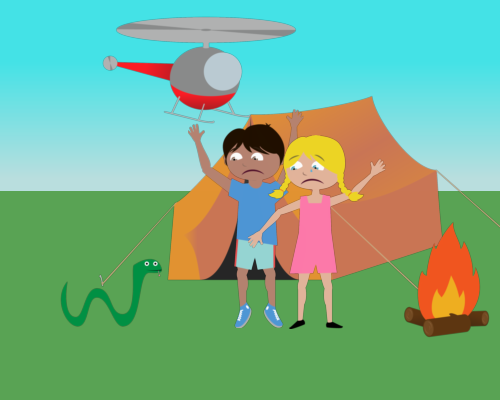} &
\includegraphics[width=0.3\linewidth]{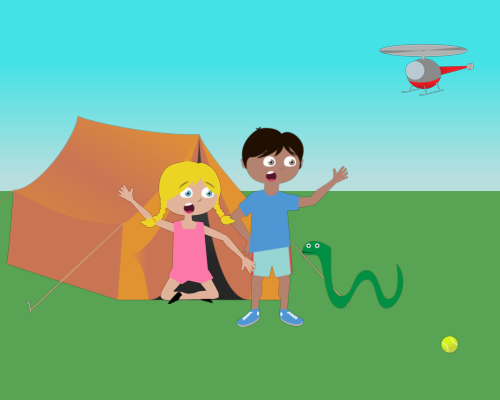} 
\end{tabular}
\caption{An example from the abstract dataset of~\cite{Zitnick_2013_CVPR}. These scenes were created from 80 clipart items by Mechanical Turk workers to illustrate the same visual scenario: ``Mike and Jenny are startled as a snake tries to get into their tent. A helicopter above is there to help.''}
\label{fig:clipart}
\end{figure}

A recent idea for learning about high-level visual concepts (e.g., interactions between objects) is to abstract away low-level visual details using non-photorealistic ``abstract art'' or cartoons. 
This is helpful as it allows researchers to safely ignore the non-semantic variability of visual categories (e.g., differences in the viewpoint, size, or orientation in which an object is portrayed, or missing parts of an object due to other objects occluding it), and to focus on the semantic differences between categories and concepts. 
\cite{Zitnick_2013_CVPR} collect a dataset of over ten thousand ``clipart'' scenes illustrating over a thousand language-based descriptions of what is happening in a scene. To create these clipart scenes, they ask annotators on  Mechanical Turk to ``create an illustration for a children's story book by creating a realistic scene from the clipart below.'' 
Annotators had 80 clipart items at their disposal. 
They were also asked to provide a written description of the scene they created, and then other annotators were asked to provide additional illustrations for the same description. Figure~\ref{fig:clipart} shows an example.
From this data, \cite{Zitnick_2013_CVPR} then learn what attributes people might have, what objects often co-occur, what spatial configurations people and objects corresponding to particular actions obey, etc. 

Since some actions may be difficult to name but easy to visualize, \cite{antol2014zero} design an interface to create cartoon depictions of actions. They crowdsource the data collection on \amt and create INTERACT, a dataset of 3,172 images that contain two people interacting, along with additional 3,000 cartoon illustrations depicting the same 60 fine-grained actions. 

Taking recognition one step further towards understanding visual content at a human-like level,
\cite{chandrasekaran2015we} collect a dataset of ``funny'' scenes, in an attempt to computationally model humor. 
They gather a dataset of over three thousand scenes by asking Mechanical Turk workers to create scenes that will be perceived by others as funny, from a richer set of clipart pieces. 
For each created scene, they also ask ten other annotators to score its degree of funniness. 
Further, they collect a separate dataset which is the ``unfunny'' counterpart to the first one. 
For each scene in the ``funny'' dataset, they ask five annotators to replace objects in the original scene in order to make it less funny. 
They verify that the resulting scenes are indeed less funny by again collecting funniness ratings.

\section{Visual story-telling}
\label{sec:visual_storytelling}

\begin{figure}
\begin{tabular}{ccc}
Events/Stories & Aesthetics & Similarity \\
\includegraphics[width=0.3\linewidth]{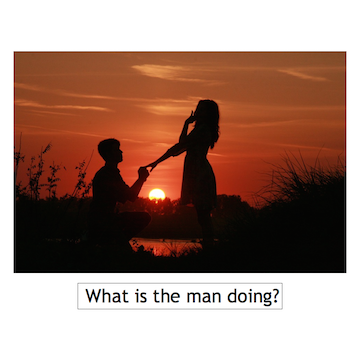} &
\includegraphics[width=0.3\linewidth]{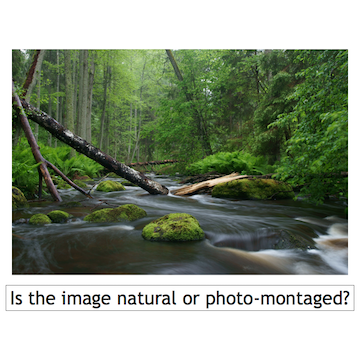} &
\includegraphics[width=0.3\linewidth]{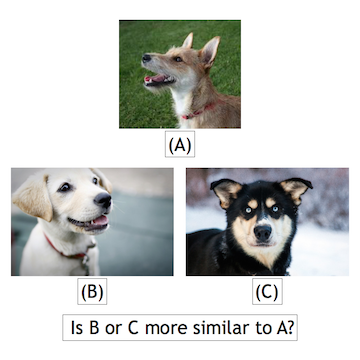}
\end{tabular}
\caption{Computer vision tasks related to visual story-telling and modeling subjective perception of the image. 
We describe strategies for collecting the corresponding annotations in Section~\ref{sec:visual_storytelling}. } 
\label{fig:buildingblocks3}
\end{figure}

We have thus far discussed how to allow a computer vision system to describe the visual content that it perceives in an image. This is done in a fairly ``documentary'' style, without any creativity or generally without room for subjectivity. 
However, images often tell entire stories, and it is useful to be able to model or replicate the creativity involved in story-telling, with computer vision techniques. In this section, we overview some initial steps to story-telling involving answering questions about images (Section \ref{sec:vqa}) and modeling subjectivity (Section \ref{sec:subj}) and perceptual similarity (Section \ref{sec:sim}). Figure~\ref{fig:buildingblocks3} summarizes these tasks.

\subsection{Visual question answering}
\label{sec:vqa}

\begin{figure}
 \centering
 \includegraphics[width=\linewidth]{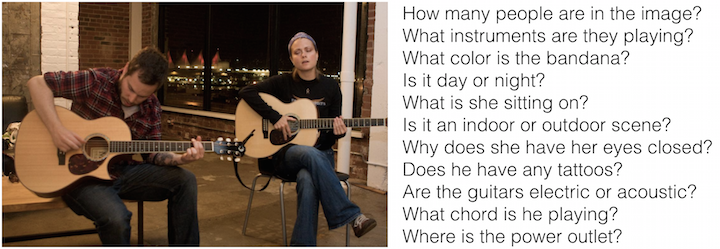}
 \caption{Some example visual questions that can be asked about an image. The expected answer can be free-form text, a region in the image, or a multiple choice selection.}
 \label{fig:vqa}
 \end{figure}

 A recent task in computer vision is visual question answering. 
 The input to the system at test time is an image and a question about this image, e.g. ``Is the person in this image expecting company?'' or ``Is the person in this image near-sighted?'' \citep{malinowski2014multi, Antol_2015_ICCV, geman2015visual}. Figure~\ref{fig:vqa} illustrates this idea with an example.
 This task is interesting because it is ``AI-complete,'' in the sense that it requires collaboration from several fields within artificial intelligence (AI). 
 In order to answer the example questions above, a computer vision systems needs to also use and represent knowledge, perform inference, and employ natural language.
 
 \cite{Antol_2015_ICCV} collect a large dataset of challenging visual questions. They show images to workers on Mechanical Turk, and ask them to write questions about the image that would be difficult for a ``smart robot'' to answer. 
 The phrasing of the task, ``Your task is to stump this smart robot,'' likely made the annotation process interesting for the workers and helped in obtaining high-quality data \citep{mao2013stop}.
 The authors separately gather ten answers per question (from ten annotators). 
 The performance of computational models for predicting answers to questions is then evaluated based on agreement with the human-given answers. 
 
 \citep{tapaswi2015movieqa} collect questions and answers about movies. 
 They encourage annotators to ask high-level questions by not showing them the movies and instead showing only text-based summaries of each movie. 
 In this particular data collection, workers were paid by the hour, supposedly motivating workers to provide higher-quality data. 
 \citep{zhu2015visual7w} collect journalism-style ``wh'' questions, ask human workers to provide pairs of questions and answers, and ask other workers to rate the quality of the question-answer pairs. 
 \citep{Yu_2015_ICCV} collect answers that can be used to fill blanks in descriptions of images.
 
 In all cases, the researchers need to collect such questions that are challenging and truly evaluate the system's ability to respond concretely to a question. Since workers might wish to optimize their monetary gain per minute, they might contribute trivial questions that either ask about unimportant visual details, or on the other extreme, do not relate tightly to the image. While the above-mentioned works use different strategies to obtain high-quality questions and answers (e.g., by engaging the user, using plot summaries, or focusing on a particular type of questions) they all aim to obtain interesting questions and accurate question-answer pairs.
 
 In the realm of a related problem, producing image descriptions,
 \citep{Vedantam_2015_CVPR} develop a new metric for evaluating automated descriptions for images, which is based on human consensus over the n-grams to be included in the answer. 
 They collect a large dataset with an impressive 50 reference descriptions (sentences) per image, since evaluating consistency is more reliable when human knowledge is captured via many samples.

\subsection{Subjective human perception of images}
\label{sec:subj}

While attributes shifted the focus of recognition towards more fine-grained descriptions of the visual world, they still focus on describing objects. 
However, there is an aspect of visual perception that goes beyond the pure physical content of images. 
For example, when looking at photographs, people might react to the style or the emotions portrayed in the image.
Several recent approaches take a first step in analyzing these aspects of the visual world.

One subjective aspect of visual content is the perceived aesthetics and artistic quality of this content. 
\cite{luo2011content} collect a database of artistic photographs from professional and amateur authors. The photographs are divided into seven content categories. The authors ask ten annotators to label each image with its quality (high or low), and a final aesthetic quality on the image is assigned if at least eight of the ten annotators provided the same label. 
\cite{simo2015neuroaesthetics} collect a dataset where the measure of clothing aesthetics is obtained from the reactions (``likes,'' ``favorites,'' etc.) on a social network. 
\cite{montagnini2012tell} collect the ``favorites'' of users on Flickr, and use this data to build models that can identify users based on their preferences. 
\cite{fan2014automated} propose a dataset used to learn whether a photograph is realistic or is heavily manipulated or synthetically constructed (``computer graphics'' or ``CG''). 
The authors collect data from about five thousand Mechanical Turk participants in total. Each image is annotated by around 30 annotators, an unusually high ``redundancy factor'' for crowdsourced data.
\cite{fan2014automated} also ask the annotators to answer 40 questions about the images, as well as to describe their background in terms of familiarity with computer games and graphic design. This allows the researchers to examine trends in how workers coming from different backgrounds label the images.

We have discussed methods that analyze \emph{what} images portray, but it is also important how viewers \emph{react} to them. 
\cite{deza2015understanding} model the popularity or ``virality'' of images, and conduct a study to determine how well humans can predict if a photograph will be popular, by having 20 annotators judge each image. 
\cite{peng2015mixed} collect a dataset of emotions evoked by images. They ask annotators to rate emotions on a scale, as well as to provide keywords that describe the evoked emotions. They apply some basic form of consistency control. 
\cite{christie2014predicting} model the user's reactions to the outputs of automated visual tasks. In particular, they wish to develop systems whose mistakes are not too \emph{annoying} to human users. To train annoyance models, they show annotators on Mechanical Turk triples of images, and ask them whether they would be more annoyed if the system returned images of the categories in B as opposed to in C, upon being queried with image A. They also ask annotators for justifications of their labels. 

A novel task just beginning to be explored in the computer vision literature is judging the implicit messages of content in visual media. 
\cite{joo2014visual} collect annotations that capture how photographs of politicians are perceived by viewers. Some photographs show the subjects in a more positive light (e.g., as ``competent,'' ``trustworthy,'' ``comforting,'' etc.) and others in a more negative light (as ``less competent,'' etc.)
The annotations are pairs of images, with only a single person portrayed in both images in the pair, and a judgement from annotators denoting which of the photographs portrays the person as having a quality to a larger extent. 
The rationale for only providing annotations on images of the same person is to avoid any personal or political bias the annotators might have. 

Collecting subjective judgements is challenging because unlike many other types of annotations, one cannot rely on a majority vote among annotators to prune noisy data. One common strategy used in the above work is to simply collect labels on the same images from many annotators, in the hope of capturing the large variability in how humans perceive and react to visual content with respect to aesthetics, emotion, etc.
Another strategy is to actively avoid annotator bias (e.g., when collecting annotations regarding the portrayal of a politician), and to focus the annotators' effort on providing data that can be used to learn how \emph{any} human would perceive some content, i.e., to build human-like visual understanding for machines.

\subsection{Perceptual and concept embeddings} 
\label{sec:sim}

 Rather than label images with having or not having some particular property, some researchers model perceptual similarity in images. 
 Unlike previous work, where the annotator was asked to say whether an image has some property, e.g. ``naturalness'' or ``aesthetics,'' here the goal is to say whether (or to what degree) two images are similar.
 \cite{tamuz2011adaptively} learn a kernel matrix that captures all pairwise similarities between objects (like one that would be used by an SVM classifier). 
 They learn this matrix from the crowd, by asking annotators which of two samples, B or C, is more similar to a query sample A. They iteratively refine their estimated kernel, from adaptively selected queries for the annotators. 
 They ensure high quality of the annotations by capping the number of tasks a user can do, and including ``test'' questions with known answers (``gold standard'' questions).
 
 \cite{Wah_2014_CVPR} also ask the user to compare similarity, by marking which of a set of images is most similar to the query image. 
 \cite{wah2015learning} request similarity comparisons on localized image patches of bird species. They first select discriminative regions, then model the probability that particular images show these regions, so they can determine which images to display to the user. 
 \cite{wilber2015learning} obtain an intuitive concept embedding by jointly optimizing an automatic low-dimensional embedding objective, as well as maximizing the probability of satisfying a set of human-given similarity triplet constraints. 

 \cite{gomes2011crowdclustering} use individual users' notions of similarity to discover object categories. Their method labels a large set of images with newly discovered categories, from user-given similarity/dissimilarity constraints (obtained from clustering). Each annotator only examines a small set of images. 
 \cite{janssens2010ranking} also integrate judgement on only a small set of images into a global judgement. 
 They ask annotators to rank a small subset of images with respect to a certain attribute, then show how to aggregate a ranking of the full set of images from different users by generating a preference matrix. 

 \cite{NIPS2015_5765} model how humans perform machine tasks, like extrapolating from function plots and finding the best fit to the data. 
 They make interesting observations: for example, humans pick the same best fit as maximum likelihood, but also are at risk of ``underfitting'' because they strive towards simpler solutions. 
 
 In all of these works, the goal is to learn how humans conceptually group visual content, so that the machine can also use a similar grouping at test time. Similarly to judging attribute presence (Section \ref{sec:attributes}), it is challenging to judge whether two images are similar, so several researchers collect data for relative similarity. Much like subjective judgements on aesthetics (Section \ref{sec:subj}), visual similarity is difficult to explain in words, so it is best captured with examples.

\section{Annotating data at different levels}
\label{sec:levels}

Usually researchers collect data that precisely matches the task they wish to ``teach'' their system.
For example, if the system's task is to predict object labels in images, researchers collect human-provided object labels on a large image dataset. 
However, researchers have shown that using an auxiliary type of data could help learn the main task. 
\cite{donahue2011annotator} collect ``annotator rationales'' which are explanations about why a certain label is present in an image. For example, if an annotator declares that a person in an image is ``attractive,'' they are also asked to draw a polygon over the image to mark which parts of the face make the person attractive. 
\cite{donahue2011annotator} then create artificial training examples with that marked part removed, and add a new SVM constraint forcing the original positive image to be scored higher for ``attractiveness'' than the image with the ``rationale'' region removed. 
The authors show that auxiliary information boosts the accuracy of classification because the system understands a little better \emph{what makes} the category present in the image, hence can capture the relevant features.

Since annotation at different levels could be helpful for different tasks, 
\cite{Branson_2014_CVPR} show how to automatically ``translate'' between different annotation types, e.g., between segmentations and object part locations.
Their translation involves a human in the loop: when the system has an estimated new annotation that it translated from another type of annotation, it presents that estimated annotation to a human for verification. 

As we showed, there is a vast number of tasks for which we need to collect annotated data from humans. Each type of data comes with its own challenges, and different (albeit related) techniques are required to ensure the quality of the data. As research in crowdsourcing for computer vision evolves, we hope the community finds a set of robust and commonly agreed-upon strategies for how to collect data. 
The next section specifically discusses approaches for data collection and quality assurance.
We also hope researchers find mechanisms through which data for different tasks can be adapted for novel tasks, so that we can make the most effective and efficient use of captured human knowledge.

\chapter{How to collect annotations}
\label{sec:how}

Having built an understanding of \emph{what} annotations we may need to collect in the computer vision community, we now turn our attention to \emph{how} these annotations should be collected. The annotation budget is always limited. Formulating an efficient and effective crowdsourcing framework can easily make the difference between being able to annotate a useful large-scale dataset that fuels computer vision research progress, and being able to only label a small handful of images. In this section, we describe insights derived both from the computer vision and the human computation literature.

\section{Interfaces for crowdsourcing and task managers}

Deploying and managing annotation tasks on a crowdsourcing platform may be a daunting job, requiring extensive UI design and backend system management to collect the results of the annotation. While \amt provides a simple framework for task management, it is often insufficient for more complex labeling tasks, such as those requiring an iterative pipeline, e.g., one worker annotates an object instance, another verifies it,  a third determines if more instances need to be annotated, and if so, the process repeats~\citep{Su12}. Further, different research groups may need similar annotation interfaces which are not always provided by \amt and other crowdsourcing platforms.

One of the first open-source efforts to standardize computer vision annotation on \amt is the  toolkit of \cite{Sorokin08}\footnote{Available at \url{http://vision.cs.uiuc.edu/annotation/}.}. It provides Flash tools and a Django web-based task management server, along with an integration with the Robotics Operating System (ROS) and annotation protocols for image segmentation and keypoint labeling. Other workflow management systems include TurKit~\citep{Little10}\footnote{Available at \url{http://groups.csail.mit.edu/uid/turkit/}.}, CLOWDER that uses decision-theoretic optimization to dynamically control the workflow~\citep{Weld11}, Turkomatic~\citep{Kulkarni12}, a cloud service tool from~\citep{Matera14}, and a recent light-weight task management system SimpleAMT\footnote{Available at \url{https://github.com/jcjohnson/simple-amt}.}.

Recent efforts in workflow management have focused on tighter feedback loops between crowd workers and the requester's goals. NEXT is a platform for facilitating active learning research that closely couples annotation with re-training machine learning models~\citep{Jamieson15}. Glance~\citep{Lasecki14} allows researchers to rapidly query, sample,
and analyze large video datasets through crowdsourcing: it temporally segments the video, distributes the annotation assignments to workers to perform in parallel, and aggregates the results within a few minutes. Glance relies on LegionTools\footnote{Available at \url{http://rochci.github.io/LegionTools/}}, an open-source framework to recruit and route workers from \amt to synchronous real-time tasks.

\begin{figure}
\begin{tabular}{cc}
\includegraphics[width=0.48\linewidth]{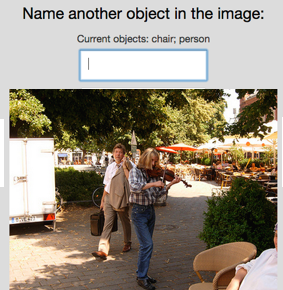} &
\includegraphics[width=0.48\linewidth]{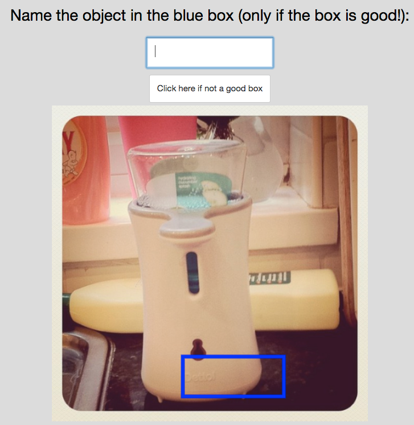} \\
\includegraphics[width=0.48\linewidth]{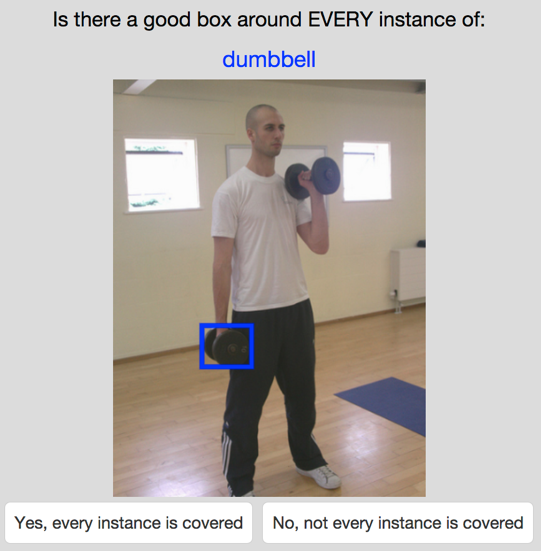} &
\includegraphics[width=0.48\linewidth]{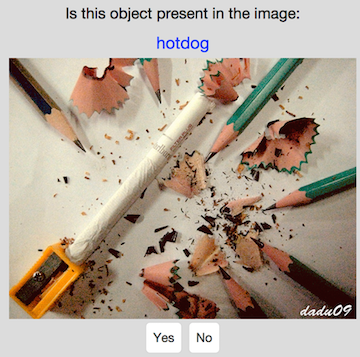} \\
\includegraphics[width=0.48\linewidth]{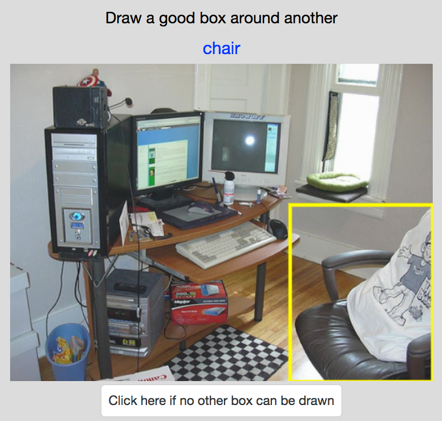} &
\includegraphics[width=0.48\linewidth]{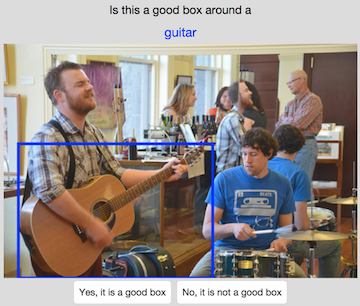} \\
\end{tabular}
\caption{Sample JavaScript interfaces for image labeling~\citep{Russakovsky15} available from \url{https://github.com/orussakovsky/annotation-UIs}.}
\label{fig:uis}
\end{figure}

Besides workflow management, several interfaces for labeling visual data are available. The LabelMe annotation tool provides an effective interface for labeling objects in images with polygons~\citep{Russell07}\footnote{Available at \url{http://labelme.csail.mit.edu/}.}. \cite{Little12} develop an interactive tool for annotating pixel-tight contours of objects. \cite{Russakovsky15} and \cite{Bearman15}\footnote{Available at \url{https://github.com/orussakovsky/annotation-UIs}.} released Javascript interfaces for multiple image labeling tasks which integrate easily with the SimpleAMT task management framework. Figure~\ref{fig:uis} illustrates some of the annotation interfaces that are available. 

In the video domain, the Janelia Automatic Animal Behavior Annotator (JAABA)~\citep{Kabra13}\footnote{Available at \url{http://jaaba.sourceforge.net/}.} provides an open-source graphical interface along with an interactive active learning backend for annotating animal behavior. VATIC~\citep{Vondrick13}\footnote{Available at \url{http://web.mit.edu/vondrick/vatic/}.} (whose use we discuss in Section \ref{sec:actions_videos}) is a popular open-source tool for labeling objects in video. It provides a framework for labeling objects with bounding boxes in a sparse set of frames, and automatically tracks them over time. iVideoSeg~\citep{Nagaraja15} is a recent toolbox for segmenting objects in video at minimal human cost. Its intuitive user interface asks annotators to provide only rough strokes rather that tight bounding boxes or precise outlines, and automatically infers the extent of the object using visual cues.

Unfortunately, often the needs of each research project are so unique that researchers end up having to design their own unique workflow management systems and/or annotation interfaces. However, these and others tools can serve as good building blocks in the design.

\section{Labeling task design}

Structuring the task to optimize labeling accuracy while minimizing worker time and effort is critical in crowdsourcing. Suboptimal design decisions can quickly lead to very costly annotation workflows.

\subsection{Effective grouping}
\label{sec:group}

Annotating a large-scale dataset can often benefit from grouping concepts together, and recent literature on cost-effective annotation has extensively explored this type of savings. \cite{Deng14} demonstrate that using a semantic hierarchy of concepts to exhaustively annotate images can yield significant time savings. Concretely, rather than asking workers individually about a concept that appears in the image, they propose asking first high-level questions such as ``is there an animal in the image?'' and only after a positive answer asking about the presence or absence of specific animals. If an image does not contain the high-level concept (e.g., an animal) then a negative answer to this single question automatically provides a negative answer to all the more specific questions (e.g., ``is there a cat?''). \cite{Chilton13} provide a way to automatically construct a hierarchical organizational strategy that can be used in this framework. Their proposed Cascade framework achieves  $80-90\%$ of the accuracy of human experts in a fraction of the time, through effective parallelization across the crowd workforce.

Using the idea of grouping concepts in a different way,~\cite{Boyko14} develop an interactive approach to labeling small objects in dense 3D LiDAR scans of a city. The system selects a group of objects, predicts a semantic label for it, and highlights it in an interactive display. The user can then confirm the label, re-label the group, or state that the objects do not belong to the same semantic class. The main technical challenge is developing an algorithm for selecting groups with many objects of the same label type, arranged in patterns that are quick for humans to recognize. 

Grouping together multiple \emph{images} into a single labeling task can be similarly effective as grouping \emph{concepts}. \cite{Wilbert14} demonstrate that collecting annotations of image similarity can be significantly more effective with a redesigned interface. Their UI shows multiple images at a time and asks workers to select two images out of six options that are most similar to a single query image, rather than existing UIs which ask workers to select the single best among two candidates. Similarly, \cite{Wigness15} avoid labeling individual images and instead solicit labels for a cluster of images. They first hierarchically cluster the images and then introduce a technique that searches for structural changes in the hierarchically clustered data as the labeling progresses. This eliminates the latency that otherwise is inherent in alternating between human labeling and image re-clustering~\citep{Biswas12,Gilbert11,Xiong12}. 

\subsection{Gamification}
\label{sec:games}

Creating a \emph{game} out of an annotation task can be a compelling way to eliminate or significantly reduce the cost of crowdsourcing. Two-player consensus-based games have been particularly popular. 
The ESP game~\citep{vonAhn05} names objects in images, Peekaboom~\citep{vonAhn06peekaboom} segments objects, Verbosity~\citep{vonAhn06verbosity} collects common-sense knowledge, ReferItGame~\citep{Kazemzadeh14} labels expressions referring to objects in images, and BubbleBank~\citep{Deng13} annotates discriminative object regions. These games usually pair crowd workers up and ask them to collaborate on a task.

For example, in the ESP game~\citep{vonAhn05} the workers are both shown the same image and each asked to name the objects they see, without seeing the other person's responses. They earn credit for any answers that match their partner's. In this way, researchers know that the object annotations are likely (a) correct, since both workers independently agree that this concept is present in the image, and (b) basic-level, since the game encourages quick and simple names that are easy for the partner to guess. 

\begin{figure}
\includegraphics[width=\linewidth]{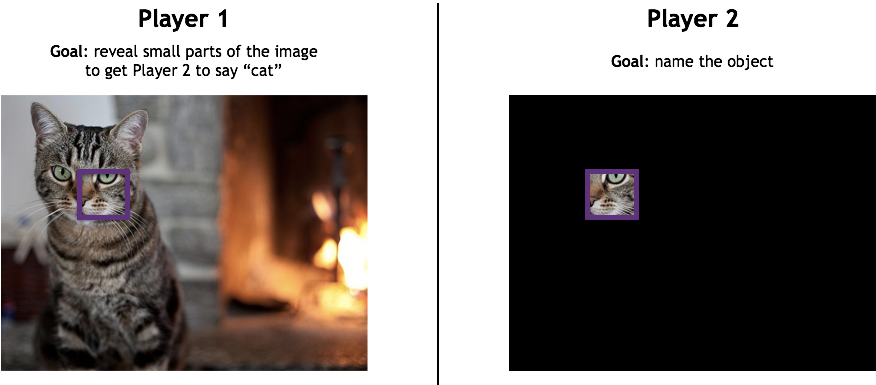}
\caption{A schematic illustration of Peekaboom~\citep{vonAhn06peekaboom}, a two-player game for annotating object locations.}
\label{fig:game}
\end{figure}

In spatial annotation games such Peekaboom~\citep{vonAhn06peekaboom},
only one worker is shown the image. This worker is also given a target visual concept in the image (such as an object of type ``cat''), and is asked to reveal small parts of the image to their partner until the partner is able to correctly guess the concept. The researchers know that the resulting spatial annotations are likely (a) correct, since the partner was able to correctly guess the target concept, and (b) locally discriminative, since the game encourages revealing as little of the image as possible. This is illustrated in Figure~\ref{fig:game}.

However, designing, deploying and maintaining a game can also be expensive. For example, large-scale annotation games such as EyeWire\footnote{Available at \url{http://eyewire.org/}.} (an online community of players who work together to segment out all neurons in the brain) is developed and maintained by professional engineers. 

\section{Evaluating and ensuring quality}
\label{sec:annotator_quality}

An important consideration when crowdsourcing annotations is ensuring the quality of the results. The three basic quality control strategies were first described by \cite{Sorokin08}: (1) build a gold standard, i.e., a collection of images with trusted annotations that are secretly injected into the task and used to verify the work; (2) design a grading task, i.e., ask workers to grade each other's work; or (3) collect multiple annotations for every input.

Strategy (1) of collecting a gold standard set requires preliminary annotation by an expert which may not always be feasible due to the added cost. In addition, the collection of a gold standard set where workers are expected to obtain perfect accuracy may not be possible when the visual recognition task is too difficult (e.g., fine-grained  classification of bird breeds). \cite{Larlus14} investigate how to design gold standard questions in this setting, such that these questions are not so easy that they are easily spotted by workers, but not so difficult that they are poor indicators of worker motivation. 

Strategy (2) of grading by other workers is particularly effective for more complex annotation, such as bounding box or image segmentation, and has been effectively utilized by \cite{Su12,ILSVRC} and \cite{COCO}. Since the original task is time-consuming and difficult to evaluate automatically, it is most effective to ask one worker to perform the task and multiple others to quickly grade the work, usually using a simple binary succeed-or-fail evaluation. Efforts such as~\citep{Russell07,Vittayakorn11} have additionally investigated automatic grading procedures to estimate annotation quality in a complex task: for example, counting the number of vertices in an annotated polygon around an object instance can serve as a proxy of segmentation quality. \cite{Shah15} propose a monetary incentive to crowd workers to only perform tasks on which they are confident, thereby encouraging self-grading.

Strategy (3), collecting multiple annotations per image, is the easiest to implement and thus has become the most popular. We focus on it for the rest of this section.

\subsection{Reconciling multiple annotators}

Asking multiple workers to annotate the same input is a proven way to obtain high-quality labels and to remove individual worker bias. We describe multiple lines of research focusing on optimally reconciling annotations obtained from multiple workers.

\cite{Sheng08} 
present repeated labeling strategies of increasing complexity, e.g., majority voting with uniform or variable worker quality estimates, round-robin strategies, and selective repeated labeling. They demonstrate that when annotators are not perfect, as expected in a crowdsourcing setting, selective acquisition of multiple labels followed by majority voting according to estimated worker quality is highly effective. Later work proposed a max-margin formulation to further improve the accuracy of majority voting~\citep{Tian15}. 

In research that tries to algorithmically evaluate the quality of the annotation work,
\cite{Welinder10cvprw,Welinder10nips,Long13active}, and \cite{Wang13quality} jointly model the labels and the annotators: i.e., they estimate the quality of each label after accumulating input from multiple annotators, and simultaneously estimate the quality of an annotator after comparing their labels to the labels provided by other workers. The method of \citep{Welinder10cvprw} is applicable generally to binary, multi-valued, and even continuous annotations such as bounding boxes. \cite{Welinder10nips} are able to discover and represent groups of annotators that have different sets of skills and knowledge, as well as groups of images that are different qualitatively. \cite{Long13active} learn a Gaussian process to derive the estimate of both the global label noise and the expertise of each individual labeler. \cite{Wang13quality} further propose quality-based pricing for crowdsourced workers after estimating the amount of information contributed by each. 

Estimating worker quality can be used not only to improve the estimation of the final label, but also for actively filtering the bad workers or selecting the best ones. Efforts such as~\citep{Hua13,Long13active} and \citep{Long15multiclass} focus on collaborative multi-annotator \emph{active learning} (discussed further in Section \ref{sec:active_learning}). They simultaneously predict the next sample to label as well as the next worker to solicit for this label, based on estimated worker quality and bias. \cite{Hua13} explore active learning with multiple oracles (perfect workers) in a collaborative setting, in contrast to most prior active learning efforts that assume a single perfect human oracle annotator. \cite{Long15multiclass} use a reinforcement learning formulation to trade off between exploration and exploitation in actively selecting both the most informative samples and the highest-quality annotators. 

Sometimes worker quality is known apriori, e.g., when having access to both expert annotators and crowd workers. \cite{Zhang15} investigate combining weak (crowd) and strong (expert) labelers. They propose an active learning algorithm which uses the weak annotators to reduce the number of label queries made to the strong annotator. \cite{Gurari15} provide practical guidance on how to collect and fuse biomedical image segmentations from experts, crowdsourced non-experts, and algorithms. \cite{Patterson15} demonstrate that just a single expert-annotated exemplar of a challenging object category (e.g., a specific breed of bird) is enough to reliably use crowd workers to annotate additional exemplars. 

In this section, we have discussed how to design the data collection process, as well as how to measure and account for different annotator expertise and bias. In the next section, we will discuss how to optimize the time of the annotator, and how to most optimally involve them in interactive learning or prediction.

\chapter{Which data to annotate} 
 \label{sec:which}

 So far, we have discussed what type of annotations to collect (i.e., ones that benefit different computer vision tasks), and how to collect them in a way that encourages quality and efficiency. 
 However, even within the same type of annotation, there are decisions to be made about which particular data instances to label with those annotations. 
 In other words, we need to \emph{actively} or \emph{interactively} select data for labeling.

 \section{Active learning}
 \label{sec:active_learning}

 Active learning is the task of deciding which data should be labeled so that the classifier learns the desired concept as fast as possible. 
 Rather than select a random batch of unlabeled data to show an annotator for labeling, an active learning system intelligently chooses the data that would be most informative to the learning system. 
 Active learning systems are usually iterative, and at each iteration, select a single sample or a batch of samples to present to a human labeler (or an oracle). 
 The selection for labeling at iteration $n+1$ is based on the classifier(s) at iteration $n$.

 \subsection{Selection criteria}
 
 \begin{figure}
     \centering
     \includegraphics[width=1\textwidth]{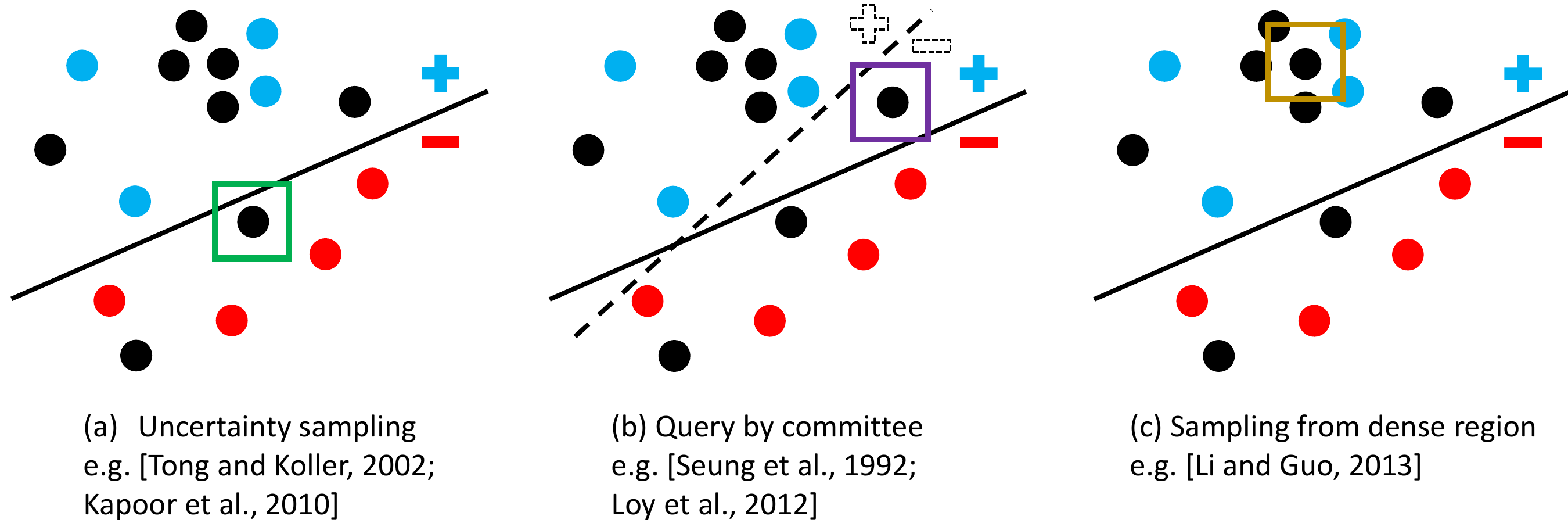}
     \caption{Three selection criteria used in active learning methods.}
     \label{fig:selection_criteria}
 \end{figure}
 
 One common criterion for selecting data to label is the uncertainty of the current classifier. 
 The system might select that sample for labeling for which the current classifier has the highest uncertainty over the class label \citep{tong2002support, kapoor2010gaussian} or has the smallest ``peak'' in the probability distribution over the class labels \citep{jain2009active}.
 Uncertainty sampling is illustrated in Figure \ref{fig:selection_criteria} (a).
 While uncertainty-based techniques are efficient, labeling an image that is uncertain does not guarantee that this label will improve certainty on all images.
 Thus, another strategy is to add all unlabeled images one at a time to the training set, and measure the expected entropy \citep{Kovashka11, kovashka2013attribute} or misclassification risk \citep{vijayanarasimhan2009s} of the updated classifier over all dataset images.
 This entropy is \emph{expected} because we do not know the true label of any image candidate for labeling, so we must weigh any entropy score by how likely it is that this image receives any particular label value \citep{Branson10}.
 
 Other active selection methods choose to have those instances labeled that lead to the largest expected model change for Gaussian processes \citep{freytag2014selecting} or conditional random fields \citep{vezhnevets2012active}, or the largest expected change in the classifier's confidence in the estimated labels \citep{wang2012active}. 
 Alternatively, the system might employ an ensemble of classifiers, and query those samples over whose labels the ensemble members most disagree \citep{seung1992query}, as illustrated in Figure \ref{fig:selection_criteria} (b). \cite{loy2012stream} apply query-by-committee to streaming data, by querying those instances that two randomly sampled hypotheses disagree over (or that at least one hypothesis places in an unknown class).
 \cite{Li_2013_CVPR} propose to request labels on images that are both uncertain and lie in a dense feature region, i.e. have high mutual information to other unlabeled samples, as illustrated in Figure \ref{fig:selection_criteria} (c).
 
  While many active selection methods are concerned with binary labeling tasks, \cite{joshi2010breaking} enable annotators to provide multi-class labels in a way that is efficient both in terms of user time (by breaking down the multi-class questions into questions with binary answers) and system selection time (developing an approximation for the selection formulation). 
 This approximation relies on making an ``optimistic'' \citep{guo2007optimistic} assumption and only considers a small set of possible label values for the unlabeled instances, instead of computing the entropy that would result from adding the unlabeled instance to the training set with every possible label value. 
 They also subsample the candidate set for labeling by clustering and only evaluating misclassification risk (which in turn is used to determine which images to label) on the set of cluster representatives. 
 
 \begin{figure}
     \centering
     \includegraphics[width=1\textwidth]{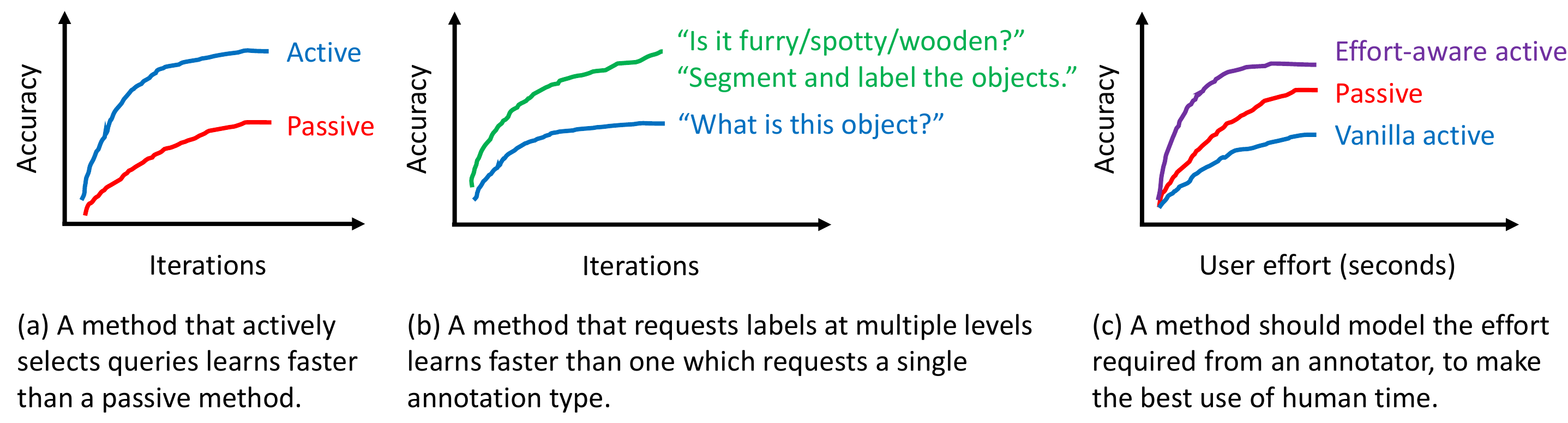}
     \caption{Three ways to measure the learning benefit that an active learning method grants.}
     \label{fig:al_curves}
 \end{figure}
 
 The goal of active selection is to make machine learning and computer vision methods learn faster the categories at hand. This is illustrated in Figure \ref{fig:al_curves} (a), where as labels are iteratively provided, the active learning method achieves higher accuracy than a passive method which does not actively solicit labels. 
 
 \subsection{Actively selecting between different types of annotations}

 As discussed in Section \ref{sec:levels}, sometimes annotations of a different type than that which the system models can be useful for the learning task. 
 If so, the system needs to select both \emph{which images} to label, as well as \emph{at what level} to label them. 
 \cite{vijayanarasimhan2009s, vijayanarasimhan2011cost} consider three types of annotations: providing object labels for the image as a whole, without labeling which region contains the object; labeling segments in an over-segmented image with an object label; or fully segmenting the image and labeling all regions. 
 They show that their active multi-level label requests allow the system to learn more efficiently than when active single-level or random label requests are used.
Efficiency is measured as the manual effort exerted by annotators. 
Similarly, \cite{Kovashka11} request labels at both the object and attribute (see Sec. \ref{sec:attributes}) levels. 
 They use \cite{wang2010discriminative}'s method which makes predictions about object labels but also models object-attribute and attribute-attribute relations. \cite{Kovashka11} show that requesting a single attribute label is more useful to the system than requesting a single object label, because attribute labels affect multiple object models. 
 The conclusion drawn in \cite{vijayanarasimhan2009s} and \cite{Kovashka11} is illustrated in Figure \ref{fig:al_curves} (b), where a method that can request among multiple types of annotations learns faster than a single-level active learning method. 
 
 \cite{vijayanarasimhan2009s, vijayanarasimhan2011cost} further model how much time it would take a user to provide any of these types of labels on any image. 
 Their method is based on timing data collected on Mechanical Turk.
 Their active selection formulation then incorporates both the risk that the current classifier might misclassify the fully labeled, partially labeled, and unlabeled data, and the cost of obtaining the labeled data. 
 Figure \ref{fig:al_curves} (c) shows that an active learning method that does not model the user effort required might underperform a passive method due to the expensive annotations requested, but if user effort is modeled, active learning retains its benefit.
 
 Several other object recognition methods also request annotations at multiple levels.
 \cite{siddiquie2010beyond} also request labels at multiple levels, and consider three types of annotations: region labels and two types of questions that capture context and 3D relationships, such as ``What is above water?'' (the answer being ``boat'') and ``What is the relationship between water and boat?'' (the answer being ``above''). 
 While they select between two label modalities in each iteration, 
 \cite{li2014multi} propose to only select within a single modality in each iteration, then adaptively choose which modality to examine. 
 \cite{parkash2012attributes} propose a new efficient attribute-based active learning approach. The system presents its predictions to the human annotator, and if the prediction is incorrect, the annotator can provide attribute-based feedback, e.g., ``This is not a giraffe because its neck is too short.'' The system then can learn that all images which have even weaker ``long neck'' attribute strength are also not giraffes. \cite{biswas2013simultaneous} use a related approach to learn object and attribute models simultaneously.
 
Beyond object recognition, active learning has also been used for 3D modeling methods.
 \cite{kowdle2011active} develop an active learning method for 3D reconstruction, by asking annotators to perform simpler tasks than previous methods require. In particular, annotators draw scribbles of different colors to mark that regions are co-planar/connected/occluding.
 Also on 3D data, \cite{Konyushkova_2015_ICCV} model uncertainty based on geometric constraints as well as the traditional feature-based uncertainty.

\subsection{Practical concerns and selecting batches of labels}

 An active learning system might request a single image or multiple images to label at the same time, i.e. perform so-called \emph{batch active learning}. 
 In the latter case, the simplest approach to select the $k$ images to label is to just sort all unlabeled images by their estimated informativeness, and take the top $k$ as the batch to label.
 However, this might result in ``myopic'' active selection, so \cite{vijayanarasimhan2010far} formulate the problem of selecting a batch to label as a ``far-sighted'' continuous optimization problem. 
 To solve this problem, they alternate between fixing the model parameters and the set of images to label. 
 In the context of relative attribute annotations and learning relative attribute models, 
 \cite{Liang_2014_CVPR} show that asking humans to fully order sets of 4 images, rather than to provide annotations on pairs of images, allows the system to learn relative attribute models faster.
 Further, the cost of obtaining the full ordering on the 4 images is about the same as on ordering just 2 images. 
 Ordering the set of 4 implicitly provides 6 ordered pairs of images, but the decision to ask for an ordering on 4 allows the annotations to be collected much more efficiently than if 6 pairs were explicitly labeled.
 
 While the benefits of many active learning techniques are demonstrated in constrained, ``sandbox'' scenarios, \cite{vijayanarasimhan2011large, vijayanarasimhan2014large} consider ``live'' active learning where they supply their method with just the name of an object category to learn, and it uses the crowd and a scalable active selection method \citep{jain2010hashing, vijayanarasimhan2014hashing} to independently learn a detector for this category.
 
 In order to model the annotation interaction between human and machine in realistic fashion,
 \cite{Kading_2015_CVPR} examine the situations when a user might refuse to provide a label because the region whose label is requested does not show a valid object, or the object cannot be recognized. Similarly,
 \cite{haines2011active} model the probability that a sample belongs to a new class. 
 
 While most of the work discussed here selects instances that will help the computer vision system \emph{learn} fast,  \cite{anirudh2014interactively} intelligently select samples to present to the user that will allow the most efficient \emph{evaluation} of the current classifier.

 \subsection{Related methods}

 Just like active learning, the goal of \emph{transfer} learning is to make learning efficient with respect to the annotation effort exerted. Transfer learning attempts to reuse \emph{existing} models in learning \emph{new} models that may be related to the existing ones. 
 \cite{Gavves_2015_ICCV} combine active and transfer learning, by considering a scenario where no data is available from a new category and proposing to use zero-shot classifiers as priors for this new unseen category. They only obtain new samples from this new category that are selected by an active learning formulation which samples from different feature space regions defined by the SVM decision boundary.

 In active learning, a human is the teacher, and the system is the learner. \cite{Johns_2015_CVPR} flip this framework around, by employing active learning to enable the system to teach a human user about difficult visual concepts. 
 Their system selects which image examples to show to the human student, and monitors the student's learning progress and ability, much like an active learning system would model what the computer has learned about visual concepts through the probability of different classes given the image. To model the human student's ability, the computer teacher shows an image only, and asks the student for a label, before revealing the true label. The goal of the teacher is to minimize the discrepancy between the student's idea of the distribution of class labels for an image, and the true distribution.
 Many ideas that \cite{Johns_2015_CVPR} use resemble active learning strategies. 
 However, as the visual concepts that computer vision study become more and more specific, such that even humans cannot easily provide the labels (e.g., breeds of dogs or types of plants and fish), machine teaching strategies will likely evolve in ways distinct from typical active learning.


 \section{Interactive annotation}

The goal of active learning is to train the most accurate computer vision model for as little annotation cost as possible. In this section, we focus on a different but related task of \emph{interactive} annotation and recognition. 
The goal is to build a collaborative human-computer system that is able to perform a given task better than either humans or computers could when working alone. Interactive systems are most useful for tasks which are apriori time-consuming or particularly difficult for the average crowd worker. 

\subsection{Interactively reducing annotation time}
\label{sec:interact_time}

Crowdsourcing at a large scale can quickly get very expensive, but the cost can be significantly reduced through the use of intelligent interactive methods.
For example, exhaustively annotating all objects with bounding boxes and class names can be a very time-consuming task, particularly on cluttered images. The goal of interactive annotation is to simplify this annotation process by utilizing computer vision models or contextual information to interactively propose object hypotheses that can then be quickly verified by humans. \cite{Yao12} present an iterative framework consisting of four steps: (1) object hypotheses are generated via a Hough Forest-based object detector, (2) the hypotheses are corrected by the user, (3) the detector is incrementally updated using the new labels, and (4) new hypotheses are generated on the fly. The authors demonstrate the effectiveness of this framework in several domains including surveillance, TV data, and cell microscopy.

Similarly, \cite{Russakovsky15} introduce a principled framework for interactively annotating objects in an image given a set of annotation constraints: desired precision (or accuracy of labeling), desired utility (loosely corresponding to number of objects annotated), and/or human cost of labeling. The annotation system incorporates seven types of human tasks, e.g., questions such as ``is there a fan in the image?'' or ``is this a bed?'' (referring to a particular bounding box). Human task selection is formulated as a Markov Decision Process, which automatically trades off between the expected increase in annotation quality and the human time required to answer the question.

Pixel-level segmentation is similarly a notoriously time-consuming task for humans. \cite{Rubinstein12} and \cite{Jain16} introduce active image segmentation systems for semantically segmenting a large set of related images. The proposed systems actively solicit human annotations for images which are likely to be most useful in propagating segmentations to other images. \cite{Nagaraja15} use motion feature cues to effectively segment objects in large-scale videos. 

\cite{Jain13} observe that while drawing detailed object segmentations is always very slow for humans, sometimes this is unnecessary. On particularly simple images, a computer vision model may be able to accurately segment out the target object given just a bounding box (which is much faster to draw) or a sloppy object contour from a human. Given image(s) that need to be segmented as quickly as possible, the proposed framework uses image features to predict the easiest annotation modality that will be sufficiently strong to yield high-quality segmentations. Extending this method, \cite{Gurari16} develop a system for automatically predicting the segmentation quality, allowing for more informed decisions about whether a computer segmentation suffices or human feedback is needed. 

Further, the time-consuming effort of annotating object parts or keypoints can be alleviated by automatically exploiting the spatial relationships between object parts. \cite{branson11strong} present a system for iterative labeling and online learning of part models, which updates and displays in real time. 

\begin{figure}
     \centering
     \includegraphics[width=1\textwidth]{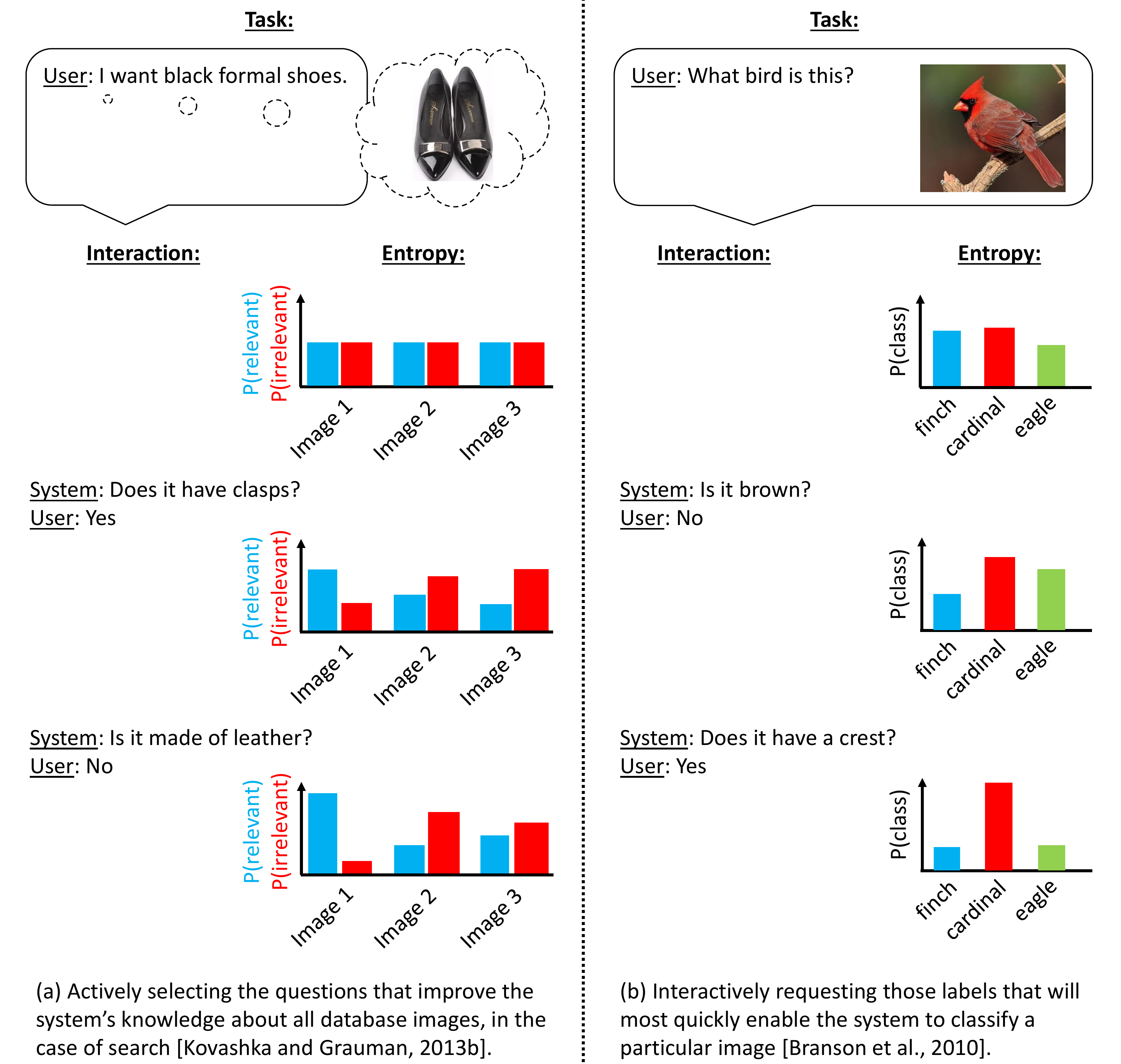}
     \caption{Two methods that use active selection.}
     \label{fig:al_questions}
 \end{figure}

 In addition to object recognition and segmentation, active selection has also been used for image retrieval. 
 \cite{kovashka2013attribute} extend \cite{kovashka2012whittlesearch}'s method for attribute-based relevance feedback, by engaging the user in a visual relative 20-questions game. The goal of the game is for the system to guess which image the user is looking for. The relative questions are composed of an image and an attribute, like ``Is the shoe you are looking for more or less \emph{shiny} than \emph{this image}?'' 
 This is illustrated in Figure \ref{fig:al_questions} (a).
 \cite{kovashka2013attribute}'s active selection method works in real time, which is necessary for an image search application. 
 To accomplish this, rather than consider all possible image-attribute questions, the systems pairs each attribute with a single image and only considers as many questions as there are attributes. The image associated with an attribute is that image which currently best exemplifies the system's guess about the desired attribute strength.
 \cite{kovashka2013attribute} demonstrate that this question-answering approach allows the user to find the image they are looking for with smaller user effort compared to providing free-form feedback as in \cite{kovashka2012whittlesearch}.
 
\subsection{Interactively improving annotation accuracy}
\label{sec:interact_accuracy}

In addition to reducing annotation time, interactive methods can also enable average crowd workers to perform tasks that are typically challenging for non-domain experts.

For example, accurate domain-specific fine-grained recognition, e.g., determining the breed of bird depicted in the image, is nearly impossible for an untrained worker. \cite{Branson10} use an interactive framework combining attribute-based human feedback with a computer vision classifier to perform this task. While humans may not be able to identify the type of bird depicted, they are able to accurately answer attribute-based questions such as ``Is the bird's belly black?'' or ``Is the bill hooked?'' The (imperfect) computer vision classifier is combined with (potentially noisy) human responses by making two assumptions: (1) human error rates are independent of image appearance, and (2) human answers are independent of each other. The method is illustrated in Figure \ref{fig:al_questions} (b).

Expanding upon this work, \cite{mensink2011learning} learn a hierarchical structure model over the attributes, to ask more informative questions and enable faster and more accurate classification. \cite{Wah11multiclass} additionally incorporate object part locations, where the user is additionally asked to click on an object part instead of answering a binary attribute question. \cite{Wah13} further extend this framework to enable zero-shot recognition, where computer vision classifiers are trained to recognize the attributes rather than the target classes. 

Attribute-based feedback has also been used for interactive clustering, where the goal is not to name the object present in the image but rather to cluster a large collection of images in a meaningful way~\citep{Lad14}. 

However, in some recognition domains such as fine-grained tree classification, it is difficult for humans not only to provide the class label but even to provide semantic attribute labels. Instead, \cite{Lee14} develop a system for tree identification that solicits humans for similarity feedback (which trees appear similar to a query tree) rather than attribute-based feedback. This feedback is used to learn a new computer vision distance metric that can then quickly recognize similar trees across multiple images. 

Such interactive methods can be effectively used to  utilize crowd workers to perform complex image annotation that otherwise would not be possible without extensive training.

\chapter{Conclusions}

In this survey, we described the computer vision tasks that have benefited from crowdsourcing annotations, i.e., from inexpensively obtaining massive datasets by distributing the labeling effort among non-expert human annotators. We described the common strategies for making the data collection efficient for both the researchers requesting the labels, and the annotators providing them. We also discussed how the quality of annotations and the skill of annotators can be evaluated, and how annotators can be encouraged to provide high-quality data. Finally, we discussed how to make the learning of computer vision models data-efficient, by intelligently selecting on which data to request labels, and by enabling vision systems to learn with interactive help from a human. 

The extensive body of literature summarized in this survey provides a solid starting block for designing a data collection strategy for a new computer vision task. Researchers have adopted some common approaches for preventing noisy data from entering the annotation pool (e.g., by including gold standard questions or reconciling labels from multiple annotators), and these approaches work reasonably well. However, with small exceptions, computer vision methods have not accounted for the large number of ``human factors'' that might affect the quality of the data that humans provide, such as attention, fatigue, miscommunication, etc. Further, researchers have not yet found a way to truly capture the vast human knowledge in a way that does not reduce this knowledge to a set of labels. We suspect that as computer vision methods become more integrated with respect to different fields of (artificial) intelligence like knowledge representation and language processing and generation, so the crowdsourcing efforts in computer vision will capture a more integrated notion of human intelligence.

\backmatter  

\bibliographystyle{abbrvnat}
\bibliography{refs_olga,refs_adriana}

\end{document}